\documentclass[10pt,journal,compsoc,twocolumn ]{IEEEtran}

\usepackage{times}
\usepackage{epsfig}
\usepackage{graphicx}
\usepackage{amsmath}
\usepackage{amssymb}
\usepackage{multirow}
\usepackage{bbm}
\usepackage{rotating}
\usepackage{array}
\usepackage{threeparttable}

\usepackage{epstopdf}

\begin{document}

\title{Deep Face Recognition: A Survey}

\author{Mei Wang, Weihong Deng
\thanks{The authors are with the Pattern Recognition and Intelligent System Laboratory, School of Information and Communication Engineering, Beijing University of Posts and Telecommunications, Beijing, 100876, China. (Corresponding Author: Weihong Deng, E-mail: whdeng@bupt.edu.cn)}}

\maketitle

\begin{abstract}

Deep learning applies multiple processing layers to learn representations of data with multiple levels of feature extraction. This emerging technique has reshaped the research landscape of face recognition (FR) since 2014, launched by the breakthroughs of DeepFace and DeepID. Since then, deep learning technique, characterized by the hierarchical architecture to stitch together pixels into invariant face representation, has dramatically improved the state-of-the-art performance and fostered successful real-world applications. In this survey, we provide a comprehensive review of the recent developments on deep FR, covering broad topics on algorithm designs, databases, protocols, and application scenes. First, we summarize different network architectures and loss functions proposed in the rapid evolution of the deep FR methods. Second, the related face processing methods are categorized into two classes: ``one-to-many augmentation'' and ``many-to-one normalization''. Then, we summarize and compare the commonly used databases for both model training and evaluation. Third, we review miscellaneous scenes in deep FR, such as cross-factor, heterogenous, multiple-media and industrial scenes. Finally, the technical challenges and several promising directions are highlighted.

\end{abstract}

\section{INTRODUCTION}

\begin{figure*}[htbp]
\centering
\includegraphics[width=\textwidth]{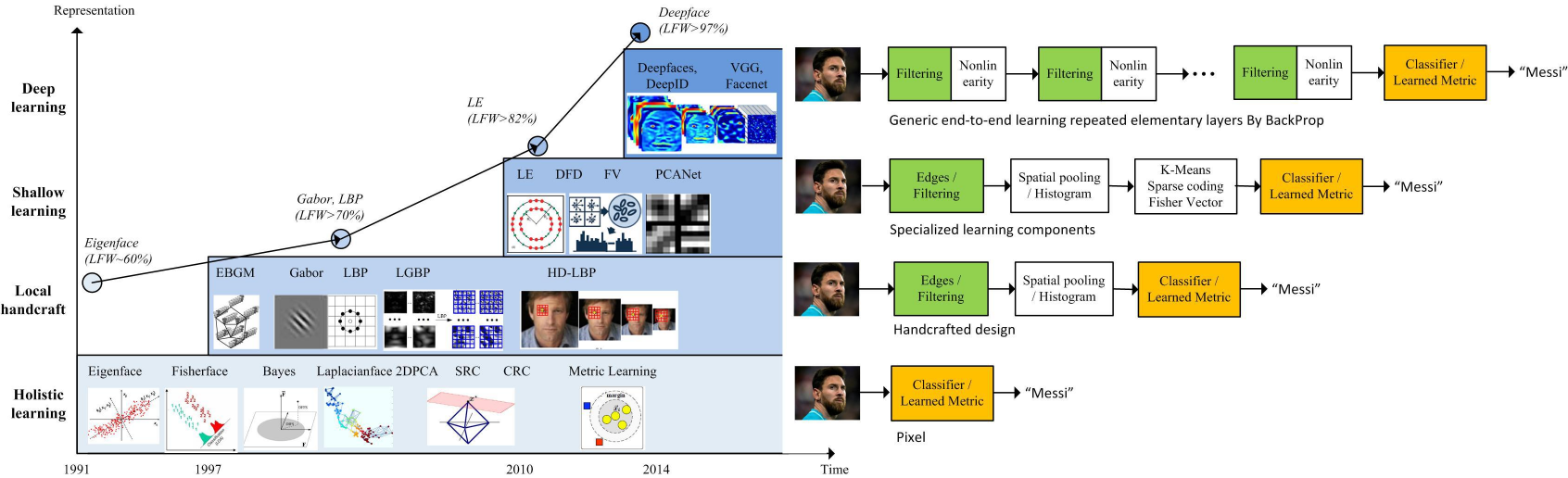}
\caption{Milestones of face representation for recognition. The holistic approaches dominated the face recognition community in the 1990s. In the early 2000s, handcrafted local descriptors became popular, and the local feature learning approaches were introduced in the late 2000s. In 2014, DeepFace \cite{taigman2014deepface} and DeepID \cite{sun2014deep} achieved a breakthrough on state-of-the-art (SOTA) performance, and research focus has shifted to deep-learning-based approaches. As the representation pipeline becomes deeper and deeper, the LFW (Labeled Face in-the-Wild) performance steadily improves from around 60\% to above 90\%, while deep learning boosts the performance to 99.80\% in just three years.}
\label{milestone}
\end{figure*}

Face recognition (FR) has been the prominent biometric technique for identity authentication and has been widely used in many areas, such as military, finance, public security and daily life. FR has been a long-standing research topic in the CVPR community. In the early 1990s, the study of FR became popular following the introduction of the historical Eigenface approach \cite{turk1991eigenfaces}. The milestones of feature-based FR over the past years are presented in Fig. \ref{milestone}, in which the times of four major technical streams are highlighted. The holistic approaches derive the low-dimensional representation through certain distribution assumptions, such as linear subspace \cite{FF}\cite{Bayesianfaces}\cite{6619386}, manifold \cite{LPP}\cite{GE}\cite{Deng-UDP}, and sparse representation \cite{wright2009robust}\cite{zhang2011sparse}\cite{deng2012extended}\cite{8053795}. This idea dominated the FR community in the 1990s and 2000s. However, a well-known problem is that these theoretically plausible holistic methods fail to address the uncontrolled facial changes that deviate from their prior assumptions. In the early 2000s, this problem gave rise to local-feature-based FR. Gabor \cite{liu2002gabor} and LBP \cite{ahonen2006face}, as well as their multilevel and high-dimensional extensions \cite{zhang2005local}\cite{chen2013blessing}\cite{8276625}, achieved robust performance through some invariant properties of local filtering. Unfortunately, handcrafted features suffered from a lack of distinctiveness and compactness. In the early 2010s, learning-based local descriptors were introduced to the FR community \cite{cao2010face}\cite{lei2014learning}\cite{chan2015pcanet}, in which local filters are learned for better distinctiveness and the encoding codebook is learned for better compactness. However, these shallow representations still have an inevitable limitation on robustness against the complex nonlinear facial appearance variations.

In general, traditional methods attempted to recognize human face by one or two layer representations, such as filtering responses, histogram of the feature codes, or distribution of the dictionary atoms. The research community studied intensively to separately improve the preprocessing, local descriptors, and feature transformation, but these approaches improved FR accuracy slowly. What's worse, most methods aimed to address one aspect of unconstrained facial changes only, such as lighting, pose, expression, or disguise. There was no any integrated technique to address these unconstrained challenges integrally. As a result, with continuous efforts of more than a decade, ``shallow'' methods only improved the accuracy of the LFW benchmark to about 95\% \cite{chen2013blessing}, which indicates that ``shallow'' methods are insufficient to extract stable identity feature invariant to real-world changes. Due to the insufficiency of this technical, facial recognition systems were often reported with unstable performance or failures with countless false alarms in real-world applications.

\begin{figure}[htbp]
\centering
\includegraphics[width=8cm]{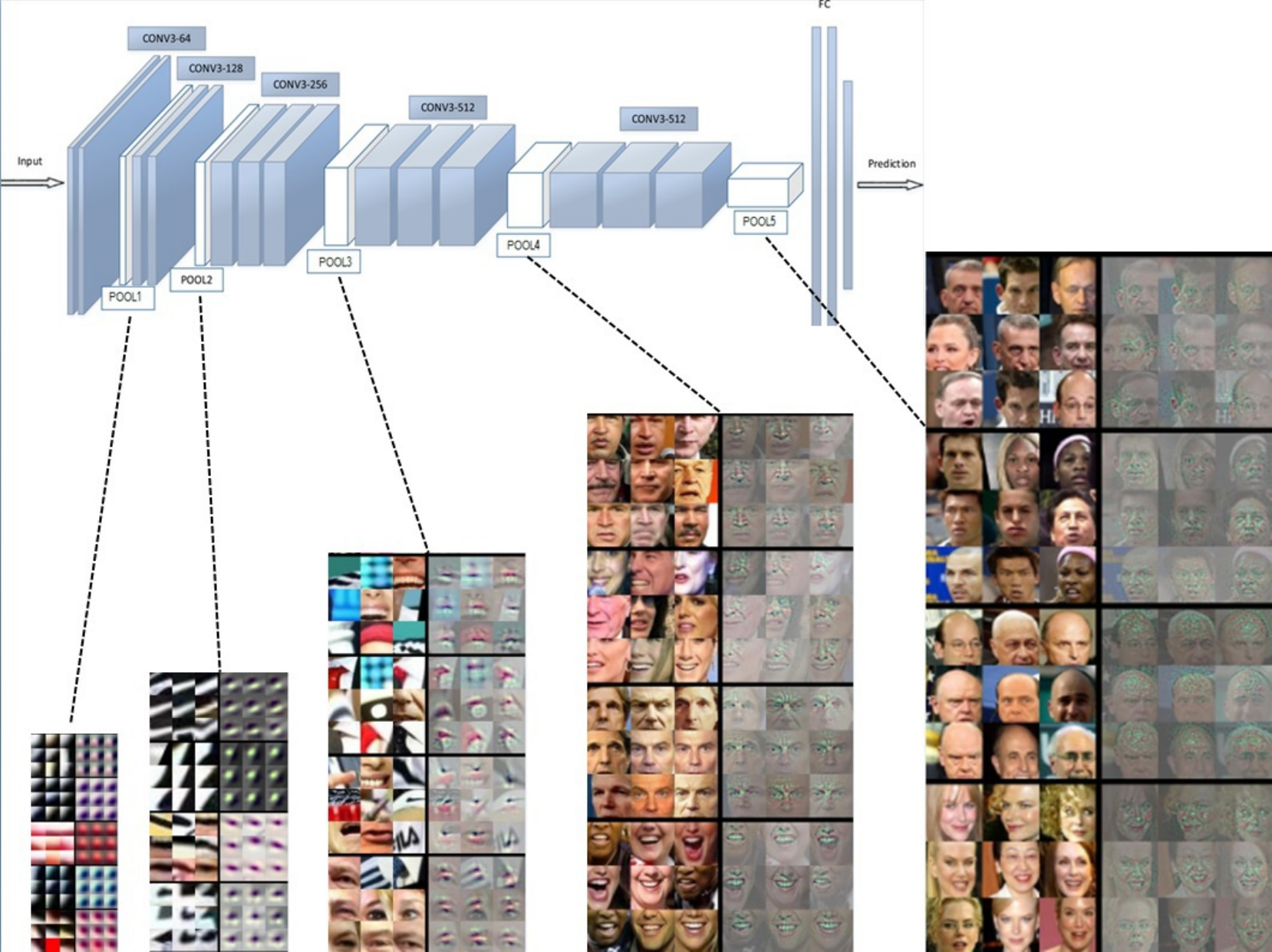}
\caption{The hierarchical architecture that stitches together pixels into invariant face representation. Deep model consists of multiple layers of simulated neurons that convolute and pool input, during which the receptive-field size of simulated neurons are continually enlarged to integrate the low-level primary elements into multifarious facial attributes, finally feeding the data forward to one or more fully connected layer at the top of the network. The output is a compressed feature vector that represent the face. Such deep representation is widely considered as the SOTA technique for face recognition.}
\label{activation}
\end{figure}

But all that changed in 2012 when AlexNet won the ImageNet competition by a large margin using a technique called deep learning \cite{krizhevsky2012imagenet}. Deep learning methods, such as convolutional neural networks, use a cascade of multiple layers of processing units for feature extraction and transformation. They learn multiple levels of representations that correspond to different levels of abstraction. The levels form a hierarchy of concepts, showing strong invariance to the face pose, lighting, and expression changes, as shown in Fig. \ref{activation}. It can be seen from the figure that the first layer of the deep neural network is somewhat similar to the Gabor feature found by human scientists with years of experience. The second layer learns more complex texture features. The features of the third layer are more complex, and some simple structures have begun to appear, such as high-bridged nose and big eyes. In the fourth, the network output is enough to explain a certain facial attribute, which can make a special response to some clear abstract concepts such as smile, roar, and even blue eye. In conclusion, in deep convolutional neural networks (CNN), the lower layers automatically learn the features similar to  Gabor and SIFT designed for years or even decades (such as initial layers in Fig. \ref{activation}), and the higher layers further learn higher level abstraction. Finally, the combination of these higher level abstraction represents facial identity with unprecedented stability.

In 2014, DeepFace \cite{taigman2014deepface} achieved the SOTA accuracy on the famous LFW benchmark \cite{huang2007labeled}, approaching human performance on the unconstrained condition for the first time (DeepFace: 97.35\% vs. Human: 97.53\%), by training a 9-layer model on 4 million facial images. Inspired by this work, research focus has shifted to deep-learning-based approaches, and the accuracy was dramatically boosted to above 99.80\% in just three years. Deep learning technique has reshaped the research landscape of FR in almost all aspects such as algorithm designs, training/test datasets, application scenarios and even the evaluation protocols. Therefore, it is of great significance to review the breakthrough and rapid development process in recent years. There have been several surveys on FR \cite{zhao2003face,bowyer2006survey,abate20072d,jafri2009survey,scheenstra2005survey} and its subdomains, and they mostly summarized and compared a diverse set of techniques related to a specific FR scene, such as illumination-invariant FR \cite{zou2007illumination}, 3D FR \cite{scheenstra2005survey}, pose-invariant FR \cite{zhang2009face}\cite{Ding2015A}. Unfortunately, due to their earlier publication dates, none of them covered the deep learning methodology that is most successful nowadays. This survey focuses only on recognition problem, and one can refer to Ranjan et al. \cite{Ranjan2018Deep} for a brief review of a full deep FR pipeline with detection and alignment, or refer to Jin et al. \cite{jin2017face} for a survey of face alignment. Specifically, the major contributions of this survey are as follows:

\begin{itemize}
\item A systematic review on the evolution of the network architectures and loss functions for deep FR is provided. Various loss functions are categorized into Euclidean-distance-based loss, angular/cosine-margin-based loss and softmax loss and its variations. Both the mainstream network architectures, such as Deepface \cite{taigman2014deepface}, DeepID series \cite{sun2014deep1,sun2015deeply,sun2014deep,sun2015deepid3}, VGGFace \cite{parkhi2015deep}, FaceNet \cite{schroff2015facenet}, and VGGFace2 \cite{cao2018vggface2}, and other architectures designed for FR are covered.
\end{itemize}

\begin{itemize}
\item We categorize the new face processing methods based on deep learning, such as those used to handle recognition difficulty on pose changes, into two classes: ``one-to-many augmentation'' and ``many-to-one normalization'', and discuss how emerging generative adversarial network (GAN) \cite{goodfellow2014generative} facilitates deep FR.
\end{itemize}

\begin{itemize}
\item We present a comparison and analysis on public available databases that are of vital importance for both model training and testing. Major FR benchmarks, such as LFW \cite{huang2007labeled}, IJB-A/B/C \cite{klare2015pushing,Whitelam2017IARPA,maze2018iarpa}, Megaface \cite{kemelmacher2016megaface}, and MS-Celeb-1M \cite{guo2016ms}, are reviewed and compared, in term of the four aspects: training methodology, evaluation tasks and metrics, and recognition scenes, which provides an useful reference for training and testing deep FR.
\end{itemize}

\begin{itemize}
\item Besides the \emph{general purpose} tasks defined by the major databases, we summarize a dozen \emph{scenario-specific} databases and solutions that are still challenging for deep learning, such as anti-attack, cross-pose FR, and cross-age FR. By reviewing specially designed methods for these unsolved problems, we attempt to reveal the important issues for future research on deep FR, such as adversarial samples, algorithm/data biases, and model interpretability.
\end{itemize}

The remainder of this survey is structured as follows. In Section II, we introduce some background concepts and terminologies, and then we briefly introduce each component of FR. In Section III, different network architectures and loss functions are presented. Then, we summarize the face processing algorithms and the datasets. In Section V, we briefly introduce several methods of deep FR used for different scenes. Finally, the conclusion of this paper and discussion of future works are presented in Section VI.

\section{OVERVIEW}

\subsection{Components of Face Recognition}

\begin{figure*}[htbp]
\centering
\includegraphics[width=\textwidth]{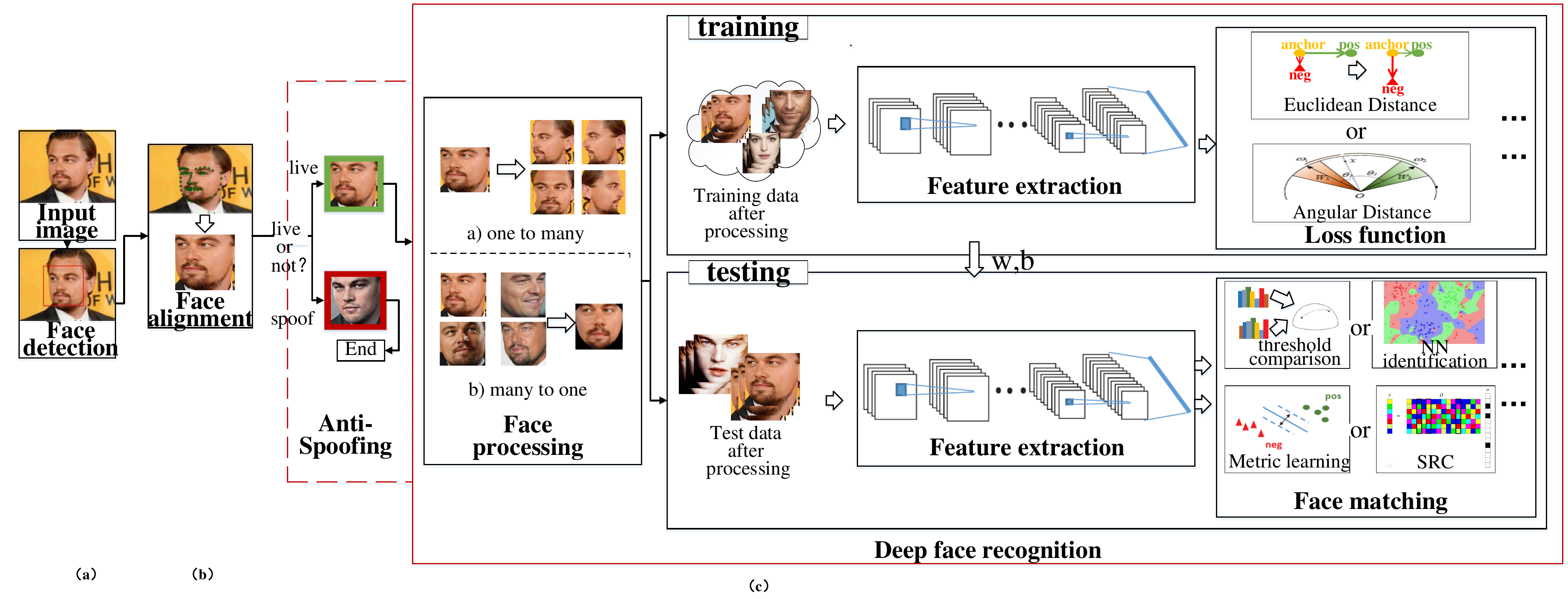}
\caption{Deep FR system with face detector and alignment. First, a face detector is used to localize faces. Second, the faces are aligned to normalized canonical coordinates. Third, the FR module is implemented. In FR module, face anti-spoofing recognizes whether the face is live or spoofed; face processing is used to handle variations before training and testing, e.g. poses, ages; different architectures and loss functions are used to extract discriminative deep feature when training; face matching methods are used to do feature classification after the deep features of testing data are extracted.}
\label{fig1}
\end{figure*}

As mentioned in \cite{Ranjan2018Deep}, there are three modules needed for FR system, as shown in Fig. \ref{fig1}. First, a face detector is used to localize faces in images or videos. Second, with the facial landmark detector, the faces are aligned to normalized canonical coordinates. Third, the FR module is implemented with these aligned face images. We only focus on the FR module throughout the remainder of this paper.

Before a face image is fed to an FR module, face anti-spoofing, which recognizes whether the face is live or spoofed, is applied to avoid different types of attacks. 
Then, recognition can be performed. As shown in Fig. \ref{fig1}(c), an FR module consists of face processing, deep feature extraction and face matching, and it can be described as follows:

\begin{equation}
M[F(P_i(I_i)),F(P_j(I_j))]
\end{equation}
where $I_i$ and $I_j$ are two face images, respectively. $P$ stands for face processing to handle intra-personal variations before training and testing, such as poses, illuminations, expressions and occlusions. $F$ denotes feature extraction, which encodes the identity information. The feature extractor is learned by loss functions when training, and is utilized to extract features of faces when testing. $M$ means a face matching algorithm used to compute similarity scores of features to determine the specific identity of faces. Different from object classification, the testing identities are usually disjoint from the training data in FR, which makes the learned classifier cannot be used to recognize testing faces. Therefore, face matching algorithm is an essential part in FR.

\subsubsection{Face Processing}

Although deep-learning-based approaches have been widely used, Mehdipour et al. \cite{mehdipour2016comprehensive} proved that various conditions, such as poses, illuminations, expressions and occlusions, still affect the performance of deep FR. Accordingly, face processing is introduced to address this problem. The face processing methods are categorized as ``one-to-many augmentation'' and ``many-to-one normalization'', as shown in Table \ref{tab2}.


\begin{itemize}
\item ``One-to-many augmentation''. These methods generate many patches or images of the pose variability from a single image to enable deep networks to learn pose-invariant representations.
\end{itemize}
\begin{itemize}
\item ``Many-to-one normalization''. These methods recover the canonical view of face images from one or many images of a nonfrontal view; then, FR can be performed as if it were under controlled conditions.
\end{itemize}
Note that we mainly focus on deep face processing method designed for pose variations in this paper, since pose is widely regarded as a major challenge in automatic FR applications and other variations can be solved by the similar methods.

\newcommand{\tabincell}[2]{\begin{tabular}{@{}#1@{}}#2\end{tabular}}  

\begin{table*}[htbp]
\small
	\centering
	\caption{Different data preprocessing approaches}
	\begin{tabular}{c|c|c}
		\hline
        \tabincell{c}{Data \\ processing}          & Brief Description             & Subsettings \\ \hline \hline
		 \multirow{2}[10]{2cm}{one to many}& \multirow{2}[10]{6cm}{These methods generate many patches or images of the pose variability from a single image} & \tabincell{c}{3D model \cite{masi2016we,masi2017rapid,richardson20163d,richardson2017learning,dou2017end}\\\cite{Guo20173DFaceNet,tuan2017regressing,tewari2017mofa}} \\ \cline{3-3}
        &    &  \tabincell{c}{2D deep model \cite{zhu2014multi,zhao2017dual,shrivastava2017learning}}\\ \cline{3-3}
         &    &  \tabincell{c}{data augmentation \cite{liu2015targeting,zhou2015naive,ding2015robust}\\ \cite{sun2015deeply,sun2014deep,sun2015deepid3,sun2016sparsifying,wang2016face}}\\ \hline
        \multirow{3}[3]{2cm}{many to one} &  \multirow{3}[0]{6cm}{These methods recover the canonical view of face images from one or many images of nonfrontal view}  & \tabincell{c}{ Antoencoder \cite{kan2014stacked,zhang2013random,yang2015weakly,zhu2013deep,zhu2014recover} } \\ \cline{3-3}
          &       & \tabincell{c}{ CNN  \cite{hu2017ldf,Zhou_2018_ECCV}}\\ \cline{3-3}
         &       & \tabincell{c}{ GAN  \cite{huang2017beyond,tran2017disentangled,deng2018uv,yin2017towards}}\\ \hline

	\end{tabular}
    \label{tab2}
\end{table*}

\subsubsection{Deep Feature Extraction}

\textbf{Network Architecture}. The architectures can be categorized as backbone and assembled networks, as shown in Table \ref{tab3}. Inspired by the extraordinary success on the ImageNet \cite{russakovsky2015imagenet} challenge, the typical CNN architectures, e.g. AlexNet, VGGNet, GoogleNet, ResNet and SENet \cite{krizhevsky2012imagenet,simonyan2014very,szegedy2015going,he2016deep,hu2018squeeze}, are introduced and widely used as the baseline models in FR (directly or slightly modified). In addition to the mainstream, some assembled networks, e.g. multi-task networks and multi-input networks, are utilized in FR. Hu et al. \cite{hu2015face} shows that accumulating the results of assembled networks provides an increase in performance compared with an individual network.

\begin{table*}[htbp]
\small
	\centering
	\caption{Different network architectures of FR}
	\begin{tabular}{c|l}
		\hline
        \tabincell{c}{Network Architectures}          & Subsettings \\ \hline \hline
		 \multirow{4}[0]{3cm}{backbone network} & \tabincell{l}{mainstream architectures: AlexNet \cite{sankaranarayanan2016triplet,sankaranarayanan2016,schroff2015facenet}, VGGNet \cite{parkhi2015deep}\\\cite{masi2016we,zhang2017range}, GoogleNet \cite{yang2017neural,schroff2015facenet}, ResNet \cite{liu2017sphereface,zhang2017range}, SENet \cite{cao2018vggface2}} \\ \cline{2-2}
        &     \tabincell{c}{light-weight architectures \cite{wu2018light,wu2015lightened,sun2016sparsifying,duong2018mobiface}}\\ \cline{2-2}
        &     \tabincell{c}{adaptive architectures \cite{zhu2020new,xiong2015conditional,Han_2018_ECCV}}\\ \cline{2-2}
        & joint alignment-recognition architectures \cite{hayat2017joint,wu2017recursive,zhong2017toward,chen2015end}\\ \hline
        assembled networks  & \tabincell{l}{ multipose \cite{kan2016multi,masi2016pose,yin2017multi,wang2014deeply}, multipatch \cite{liu2015targeting,zhou2015naive,ding2015robust,sun2013hybrid,sun2014deep1,sun2014deep}\\ \cite{sun2015deeply}, multitask \cite{ranjan2017all} } \\ \hline

	\end{tabular}
    \label{tab3}
\end{table*}

\textbf{Loss Function}. The softmax loss is commonly used as the supervision signal in object recognition, and it encourages the separability of features. However, the softmax loss is not sufficiently effective for FR because intra-variations could be larger than inter-differences and more discriminative features are required when recognizing different people. Many works focus on creating novel loss functions to make features not only more separable but also discriminative, as shown in Table \ref{tab4}.

\begin{table*}[htbp]
\small
\centering
\caption{Different loss functions for FR}
\setlength{\tabcolsep}{1mm}{
 \begin{tabular}{c|l}
  \hline
   \tabincell{c}{Loss Functions} & Brief Description \\ \hline \hline
   \tabincell{c}{Euclidean-distance-\\based loss} & \tabincell{l}{These methods reduce intra-variance and enlarge inter-variance based on \\Euclidean distance. \cite{sun2014deep,sun2015deeply,sun2015deepid3,wen2016discriminative,wu2017deep,zhang2017range,schroff2015facenet,parkhi2015deep,sankaranarayanan2016triplet,sankaranarayanan2016,
   liu2015targeting,chen2016unconstrained}} \\ \hline
   \tabincell{c}{angular/cosine-margin-\\based loss} & \tabincell{l}{These methods make learned features potentially separable with larger \\angular/cosine distance. \cite{liu2016large,liu2017sphereface,wang2018additive,deng2019arcface,wang2018cosface,liu2017deep}} \\ \hline
   \tabincell{c}{softmax loss and its \\variations} & \tabincell{l}{These methods modify the softmax loss to improve performance, e.g. \\features or weights normalization. \cite{ranjan2017l2,wang2017normface,hasnat2017deepvisage,liu2017rethinking,qi2018face,chen2017noisy,hasnat2017mises}} \\ \hline
 \end{tabular}}
 \label{tab4}
\end{table*}




\subsubsection{Face Matching by Deep Features}

FR can be categorized as face verification and face identification. In either scenario, a set of known subjects is initially enrolled in the system (the gallery), and during testing, a new subject (the probe) is presented. After the deep networks are trained on massive data with the supervision of an appropriate loss function, each of the test images is passed through the networks to obtain a deep feature representation. Using cosine distance or L2 distance, face verification computes one-to-one similarity between the gallery and probe to determine whether the two images are of the same subject, whereas face identification computes one-to-many similarity to determine the specific identity of a probe face. 
In addition to these, other methods are introduced to postprocess the deep features such that the face matching is performed efficiently and accurately, such as metric learning, sparse-representation-based classifier (SRC), and so forth.

\begin{figure}[htbp]
\centering
\includegraphics[width=8cm]{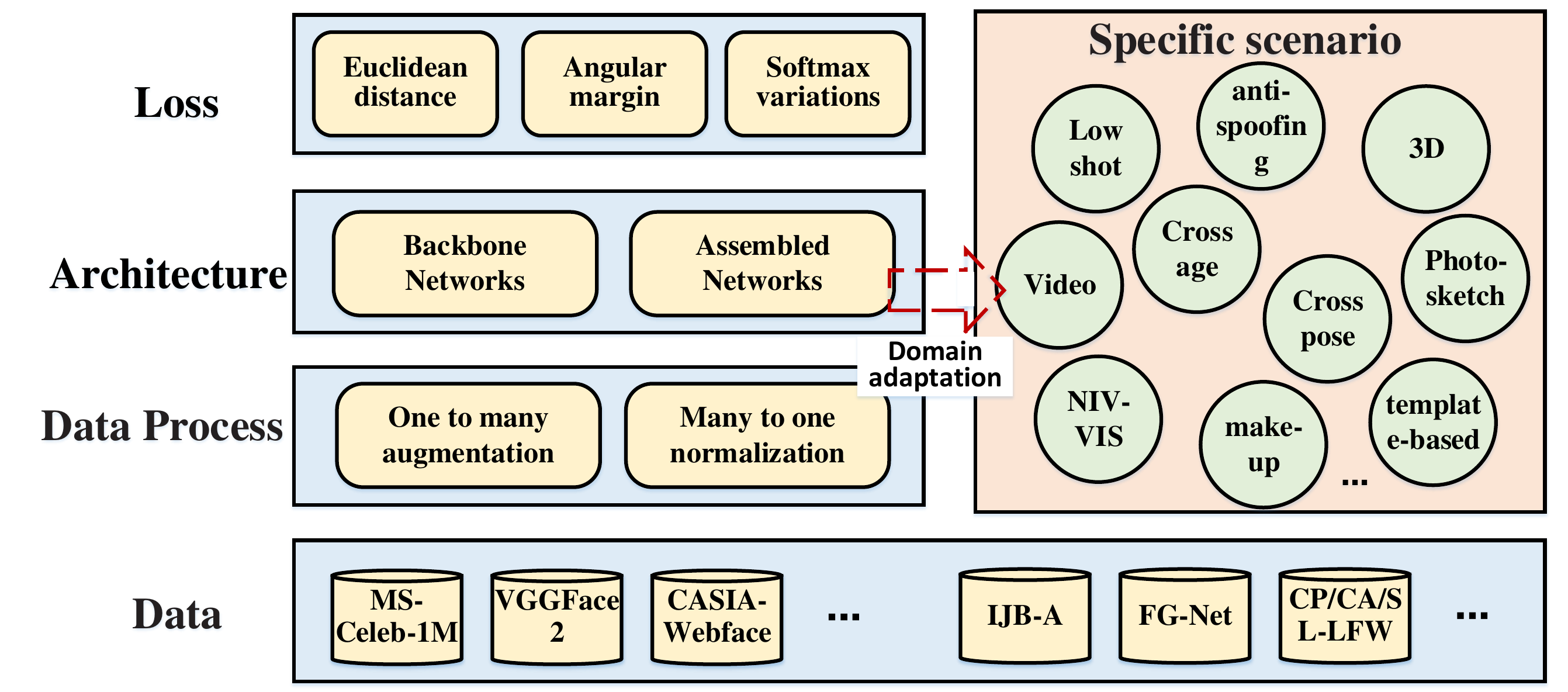}
\caption{ FR studies have begun with general scenario, then gradually get close to more realistic applications and drive different solutions for specific scenarios, such as cross-pose FR, cross-age FR, video FR. In specific scenarios, targeted training and testing database are constructed, and face processing, architectures and loss functions are modified based on the special requirements.}
\label{fig19}
\end{figure}

To sum up, we present FR modules and their commonly-used methods in Fig. \ref{fig19} to help readers to get a view of the whole FR. In deep FR, various training and testing face databases are constructed, and different architectures and losses of deep FR always follow those of deep object classification and are modified according to unique characteristics of FR. Moreover, in order to address unconstrained facial changes, face processing methods are further designed to handle poses, expressions and occlusions variations. Benefiting from these strategies, deep FR system significantly improves the SOTA and surpasses human performance. When the applications of FR becomes more and more mature in general scenario, recently, different solutions are driven for more difficult specific scenarios, such as cross-pose FR, cross-age FR, video FR. 

\section{Network Architecture and Training Loss}

For most applications, it is difficult to include the candidate faces during the training stage, which makes FR become a ``zero-shot'' learning task. Fortunately, since all human faces share a similar shape and texture, the representation learned from a small proportion of faces can generalize well to the rest. Based on this theory, a straightforward way to improve generalized performance is to include as many IDs as possible in the training set. For example, Internet giants such as Facebook and Google have reported their deep FR system trained by $10^{6}-10^{7}$ IDs \cite{schroff2015facenet,taigman2014deepface}.

Unfortunately, these personal datasets, as well as prerequisite GPU clusters for distributed model training, are not accessible for academic community. Currently, public available training databases for academic research consist of only $10^{3}-10^{5}$ IDs. Instead, academic community makes effort to design effective loss functions and adopts efficient architectures to make deep features more discriminative using the relatively small training data sets. For instance, the accuracy of most popular LFW benchmark has been boosted from 97\% to above 99.8\% in the pasting four years, as enumerated in Table \ref{tab10}. In this section, we survey the research efforts on different loss functions and network architectures that have significantly improved deep FR methods.

\begin{sidewaystable*}[htbp]
\centering
\begin{threeparttable}
\caption{The accuracy of different methods evaluated on the LFW dataset.}
\setlength{\tabcolsep}{0.5mm}{
 \begin{tabular}{c|c|c|c|c|c|c}
  \hline
   Method                                 & \tabincell{l}{Public.\\ Time} & Loss      & Architecture       & \tabincell{l}{Number of \\Networks}  & Training Set & Accuracy$\pm$Std(\%) 
   \\ \hline \hline
   DeepFace \cite{taigman2014deepface}    &2014 & softmax   & Alexnet            & 3                   & Facebook (4.4M,4K)  & 97.35$\pm$0.25      
   \\ \hline
   DeepID2 \cite{sun2014deep}             &2014& contrastive loss & Alexnet     & 25                  & CelebFaces+ (0.2M,10K) & 99.15$\pm$0.13    
   \\ \hline
   DeepID3 \cite{sun2015deepid3}          &2015& contrastive loss & VGGNet-10   & 50                  & CelebFaces+ (0.2M,10K) & 99.53$\pm$0.10    
   \\ \hline
   FaceNet \cite{schroff2015facenet}      &2015& triplet loss     & GoogleNet-24 & 1                  & Google (500M,10M)      & 99.63$\pm$0.09    
   \\ \hline
   Baidu \cite{liu2015targeting}          &2015& triplet loss     & CNN-9        & 10                  & Baidu (1.2M,18K)       & 99.77             
   \\ \hline
   VGGface \cite{parkhi2015deep}          &2015& triplet loss     & VGGNet-16    & 1                  & VGGface (2.6M,2.6K)      & 98.95            
   \\ \hline
   light-CNN \cite{wu2018light}            &2015& softmax         & light CNN    & 1                  & MS-Celeb-1M (8.4M,100K)   & 98.8            
   \\ \hline
   Center Loss \cite{wen2016discriminative} &2016& center loss    & Lenet+-7     & 1                  & \tabincell{l}{CASIA-WebFace, CACD2000,\\ Celebrity+ (0.7M,17K)} & 99.28 
   \\ \hline
   L-softmax \cite{liu2016large}           &2016& L-softmax       & VGGNet-18    & 1                  & CASIA-WebFace (0.49M,10K) & 98.71           
   \\ \hline
   Range Loss \cite{zhang2017range}        &2016& range loss       & VGGNet-16    & 1                 & \tabincell{l}{MS-Celeb-1M, CASIA-WebFace\\ (5M,100K)} & 99.52 
   \\ \hline
   L2-softmax \cite{ranjan2017l2}          &2017& L2-softmax      & ResNet-101    &1                  & MS-Celeb-1M (3.7M,58K)    &99.78            
   \\ \hline
   Normface \cite{wang2017normface}        &2017& contrastive loss & ResNet-28   & 1                  & CASIA-WebFace (0.49M,10K) & 99.19           
   \\ \hline
   CoCo loss \cite{liu2017rethinking}      &2017& CoCo loss        & -           &1                   & MS-Celeb-1M (3M,80K)      & 99.86           
   \\ \hline
   vMF loss \cite{hasnat2017mises}         &2017& vMF loss        & ResNet-27    & 1                  & MS-Celeb-1M (4.6M,60K)    & 99.58           
   \\ \hline
   Marginal Loss \cite{deng2017marginal}    &2017& marginal loss    & ResNet-27    & 1                 & MS-Celeb-1M (4M,80K)       & 99.48          
   \\ \hline
   SphereFace \cite{liu2017sphereface}     &2017& A-softmax        & ResNet-64    &1                  & CASIA-WebFace (0.49M,10K)   & 99.42         
   \\ \hline
   CCL \cite{qi2018face}                   &2018& center invariant loss & ResNet-27 & 1               & CASIA-WebFace (0.49M,10K) & 99.12           
   \\ \hline
   AMS loss \cite{wang2018additive}        &2018& AMS loss         & ResNet-20    & 1                  & CASIA-WebFace (0.49M,10K)  & 99.12         \\ \hline
   Cosface \cite{wang2018cosface}          &2018& cosface          & ResNet-64    & 1                  & CASIA-WebFace (0.49M,10K)  & 99.33 \\ \hline
   Arcface \cite{deng2019arcface}          &2018& arcface          & ResNet-100   & 1                  & MS-Celeb-1M (3.8M,85K)     &99.83\\ \hline
   Ring loss \cite{Zheng_2018_CVPR}        &2018& Ring loss        & ResNet-64    & 1                  & MS-Celeb-1M (3.5M,31K)     & 99.50\\ \hline
 \end{tabular}}
 \label{tab10}
 \end{threeparttable}
\end{sidewaystable*}

\subsection{Evolution of Discriminative Loss Functions}

\begin{figure*}[htbp]
\centering
\includegraphics[width=\textwidth]{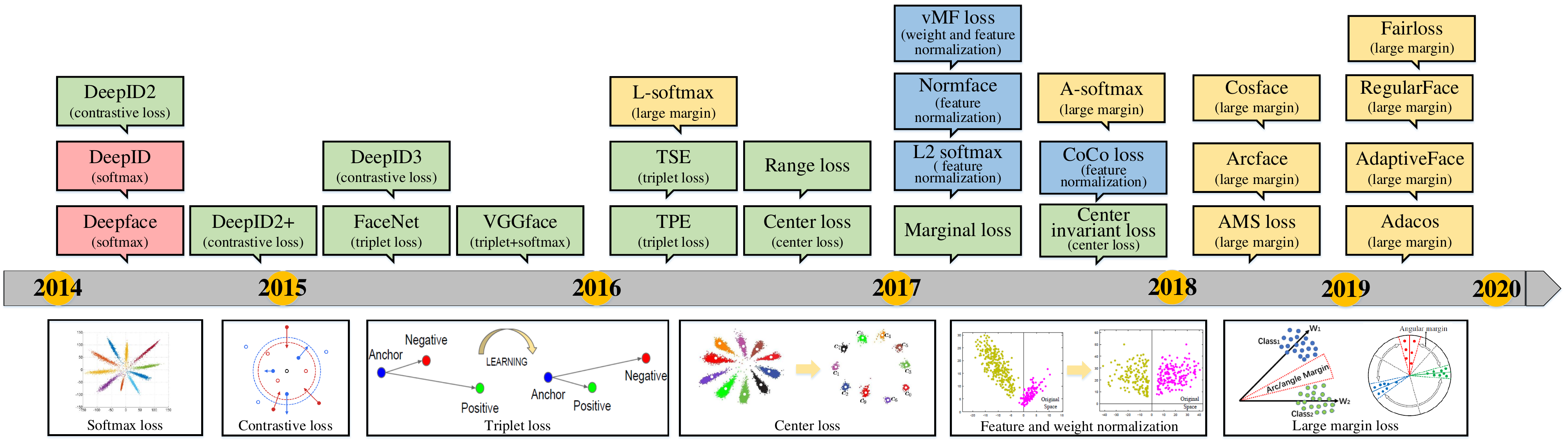}
\caption{ The development of loss functions. It marks the beginning of deep FR that Deepface \cite{taigman2014deepface} and DeepID \cite{sun2014deep1} were introduced in 2014. After that, Euclidean-distance-based loss always played the important role in loss function, such as contractive loss, triplet loss and center loss. In 2016 and 2017, L-softmax \cite{liu2016large} and A-softmax \cite{liu2017sphereface} further promoted the development of the large-margin feature learning. In 2017, feature and weight normalization also begun to show excellent performance, which leads to the study on variations of softmax. Red, green, blue and yellow rectangles represent deep methods using softmax, Euclidean-distance-based loss, angular/cosine-margin-based loss and variations of softmax, respectively. }
\label{fig21}
\end{figure*}

Inheriting from the object classification network such as AlexNet,  the initial Deepface \cite{taigman2014deepface} and DeepID \cite{sun2014deep1} adopted cross-entropy based softmax loss for feature learning. After that, people realized that the softmax loss is not sufficient by itself to learn discriminative features, and more researchers began to explore novel loss functions for enhanced generalization ability. This becomes the hottest research topic in deep FR research, as illustrated in Fig. \ref{fig21}. Before 2017, Euclidean-distance-based loss played an important role; In 2017, angular/cosine-margin-based loss as well as feature and weight normalization became popular. It should be noted that, although some loss functions share the similar basic idea, the new one is usually designed to facilitate the training procedure by easier parameter or sample selection.

\subsubsection{ Euclidean-distance-based Loss }

Euclidean-distance-based loss is a metric learning method \cite{xing2003distance,weinberger2009distance} that embeds images into Euclidean space in which intra-variance is reduced and inter-variance is enlarged. The contrastive loss and the triplet loss are the commonly used loss functions. The contrastive loss \cite{sun2015deeply,sun2014deep,sun2015deepid3,sun2016sparsifying,yi2014learning} requires face image pairs, and then pulls together positive pairs and pushes apart negative pairs.
\begin{equation}
\begin{split}
\mathcal{L}=&y_{ij}max\left ( 0,\left \| f(x_{i})-f(x_{j}) \right \|_{2}-\epsilon ^{+} \right )\\&+(1-y_{ij})max\left (  0,\epsilon ^{-}-\left \|  f(x_{i})-f(x_{j}) \right \|_{2}\right )
\end{split}
\end{equation}
where $y_{ij}=1$ means $x_{i}$ and $x_{j}$ are matching samples and $y_{ij}=0$ means non-matching samples. $f(\cdot)$ is the feature embedding, $\epsilon^{+}$ and $\epsilon^{-}$ control the margins of the matching and non-matching pairs respectively. DeepID2 \cite{sun2014deep} combined the face identification (softmax) and verification (contrastive loss) supervisory signals to learn a discriminative representation, and joint Bayesian (JB) was applied to obtain a robust embedding space. Extending from DeepID2 \cite{sun2014deep}, DeepID2+ \cite{sun2015deeply} increased the dimension of hidden representations and added supervision to early convolutional layers. DeepID3 \cite{sun2015deepid3} further introduced VGGNet and GoogleNet to their work. However, the main problem with the contrastive loss is that the margin parameters are often difficult to choose.

Contrary to contrastive loss that considers the absolute distances of the matching pairs and non-matching pairs, triplet loss considers the relative difference of the distances between them. Along with FaceNet \cite{schroff2015facenet} proposed by Google, Triplet loss \cite{schroff2015facenet,parkhi2015deep,sankaranarayanan2016,sankaranarayanan2016triplet,liu2015targeting,ding2015robust} was introduced into FR. It requires the face triplets, and then it minimizes the distance between an anchor and a positive sample of the same identity and maximizes the distance between the anchor and a negative sample of a different identity. FaceNet made $\left \| f(x_{i}^{a})-f(x_{i}^{p})\right \|_{2}^{2}+\alpha<  - \left \| f(x_{i}^{a})-f(x_{i}^{n})\right \|_{2}^{2} $ using hard triplet face samples, where $x_{i}^{a}$, $x_{i}^{p}$ and $x_{i}^{n}$ are the anchor, positive and negative samples, respectively, $\alpha$ is a margin and $f(\cdot)$ represents a nonlinear transformation embedding an image into a feature space. Inspired by FaceNet \cite{schroff2015facenet}, TPE \cite{sankaranarayanan2016} and TSE \cite{sankaranarayanan2016triplet} learned a linear projection $W$ to construct triplet loss. 
Other methods optimize deep models using both triplet loss and softmax loss \cite{zhou2015naive,liu2015targeting,ding2015robust,crosswhite2017template}. They first train networks with softmax and then fine-tune them with triplet loss.

However, the contrastive loss and triplet loss occasionally encounter training instability due to the selection of effective training samples, some paper begun to explore simple alternatives. Center loss \cite{wen2016discriminative} and its variants \cite{zhang2017range,deng2017marginal,wu2017deep} are good choices for reducing intra-variance. The center loss \cite{wen2016discriminative} learned a center for each class and penalized the distances between the deep features and their corresponding class centers. This loss can be defined as follows:
\begin{equation}
\mathcal{L}_{C}=\frac{1}{2}\sum_{i=1}^{m}\left \|  x_{i}-c_{y_{i}}\right \|_{2}^{2}
\end{equation}
where $x_i$ denotes the $i$-th deep feature belonging to the $y_i$-th class and $c_{y_{i}}$ denotes the $y_i$-th class center of deep features. To handle the long-tailed data, a range loss \cite{zhang2017range}, which is a variant of center loss, is used to minimize k greatest range's harmonic mean values in one class and maximize the shortest inter-class distance within one batch. Wu et al. \cite{wu2017deep} proposed a center-invariant loss that penalizes the difference between each center of classes. Deng et al. \cite{deng2017marginal} selected the farthest intra-class samples and the nearest inter-class samples to compute a margin loss. However, the center loss and its variants suffer from massive GPU memory consumption on the classification layer, and prefer balanced and sufficient training data for each identity.

\subsubsection{ Angular/cosine-margin-based Loss }

\begin{figure}[htbp]
\centering
\includegraphics[width=8cm]{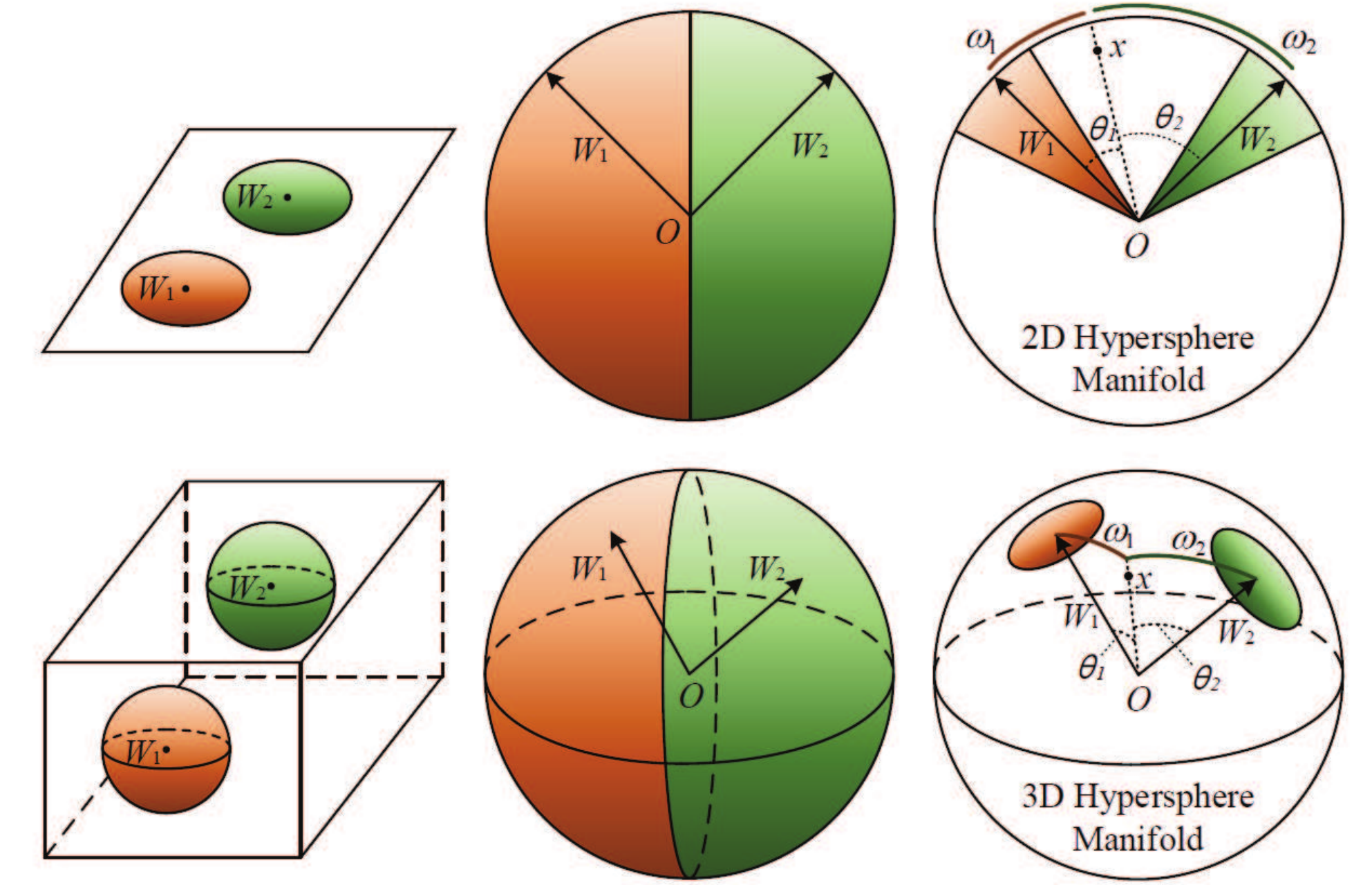}
\caption{ Geometry interpretation of A-Softmax loss. \cite{liu2017sphereface}}
\label{fig9}
\end{figure}

\begin{figure}[htbp]
\centering
\includegraphics[width=8cm]{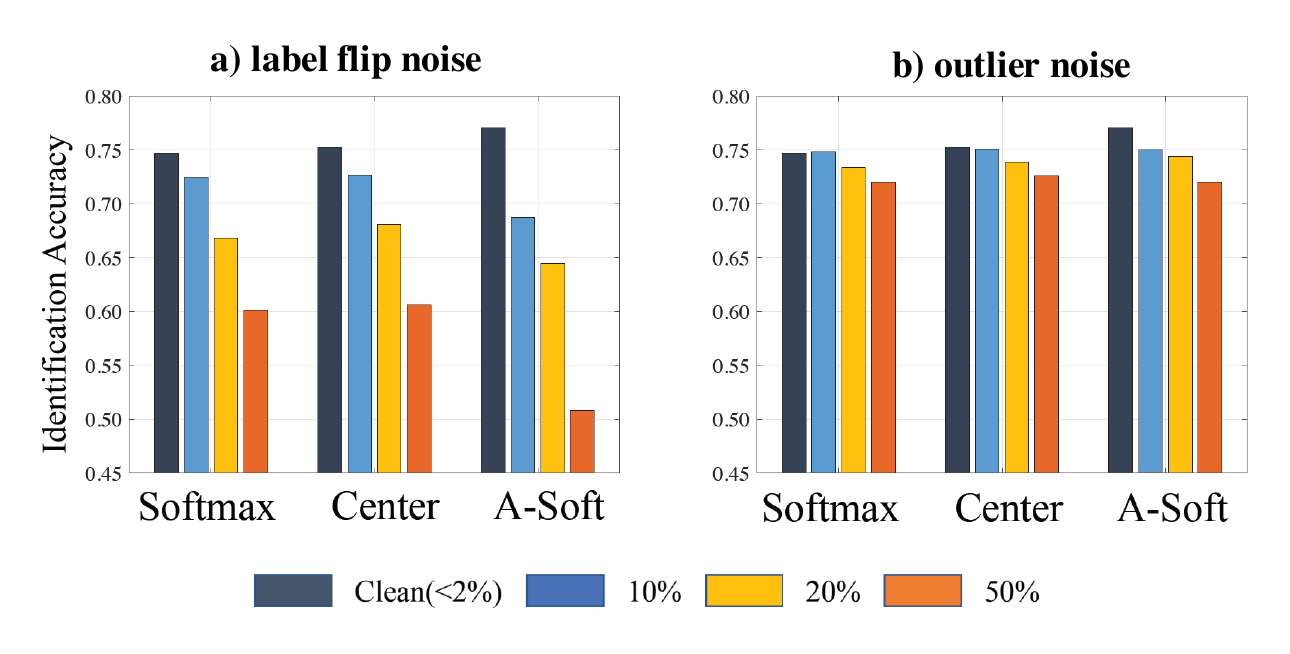}
\caption{ 1:1M rank-1 identification results on MegaFace benchmark: (a) introducing label flips to training data, (b) introducing outliers to training data. \cite{Wang_2018_ECCV}}
\label{fig26}
\end{figure}

In 2017, people had a deeper understanding of loss function in deep FR and thought that samples should be separated more strictly to avoid misclassifying the difficult samples.  Angular/cosine-margin-based loss \cite{liu2016large,liu2017sphereface,wang2018additive,deng2019arcface,liu2017deep} is proposed to make learned features potentially separable with a larger angular/cosine distance. The decision boundary in softmax loss is $\left ( W_1-W_2 \right )x+b_1-b_2=0$, where $x$ is feature vector, $W_i$ and $b_i$ are weights and bias in softmax loss, respectively. Liu et al. \cite{liu2016large} reformulated the original softmax loss into a large-margin softmax (L-Softmax) loss. They constrain $b_1=b_2 =0$, so the decision boundaries for class 1 and class 2 become $\left \| x \right \|\left (\left \| W_1 \right \| cos\left ( m\theta _1 \right )-\left \| W_2 \right \|cos\left ( \theta _2 \right ) \right )=0$ and $\left \| x \right \|\left ( \left \| W_1 \right \|\left \| W_2 \right \|cos\left ( \theta _1 \right )-cos\left ( m\theta _2 \right ) \right )=0$, respectively, where $m$ is a positive integer introducing an angular margin, and $\theta_i$ is the angle between $W_i$ and $x$. Due to the non-monotonicity of the cosine function, a piece-wise function is applied in L-softmax to guarantee the monotonicity. The loss function is defined as follows:
 \begin{equation}
\mathcal{L}_{i}=-log\left ( \frac{e^{\left \| W_{yi} \right \|\left \| x_{i} \right \|\varphi (\theta _{yi})}}{e^{\left \| W_{yi} \right \|\left \| x_{i} \right \|\varphi (\theta _{yi})+\sum _{j\neq y_{i}}e^{\left \| W_{yi} \right \|\left \| x_{i} \right \|cos(\theta _{j})} }}\right )
\end{equation}
where
 \begin{equation}
\varphi (\theta )=(-1)^{k}cos(m\theta )-2k,\theta \in\left [  \frac{k\pi }{m},\frac{(k+1)\pi }{m}\right ]
\end{equation}
Considering that L-Softmax is difficult to converge, it is always combined with softmax loss to facilitate and ensure the convergence. Therefore, the loss function is changed into: $f_{y_{i}}=\frac{\lambda\left \| W_{y_{i}} \right \|\left \| x_{i} \right \|cos(\theta _{y_{i}})+\left \| W_{y_{i}} \right \|\left \| x_{i} \right \|\varphi (\theta _{y_{i}})}{1+\lambda }$, where $\lambda$ is a dynamic hyper-parameter. Based on L-Softmax, A-Softmax loss \cite{liu2017sphereface} further normalized the weight $W$ by L2 norm ($\left \| W \right \|=1$) such that the normalized vector will lie on a hypersphere, and then the discriminative face features can be learned on a hypersphere manifold with an angular margin (Fig. \ref{fig9}). Liu et al. \cite{liu2017deep} introduced a deep hyperspherical convolution network (SphereNet) that adopts hyperspherical convolution as its basic convolution operator and is supervised by angular-margin-based loss. To overcome the optimization difficulty of L-Softmax and A-Softmax, which incorporate the angular margin in a multiplicative manner, ArcFace \cite{deng2019arcface} and CosFace \cite{wang2018additive}, AMS loss \cite{wang2018cosface} respectively introduced an additive angular/cosine margin $cos(\theta+m)$ and $cos\theta-m$. They are extremely easy to implement without tricky hyper-parameters $\lambda$, and are more clear and able to converge without the softmax supervision. The decision boundaries under the binary classification case are given in Table \ref{tab6}. Based on large margin, FairLoss \cite{Liu_2019_ICCV} and AdaptiveFace \cite{Liu_2019_CVPR} further proposed to adjust the margins for different classes adaptively to address the problem of unbalanced data. Compared to Euclidean-distance-based loss, angular/cosine-margin-based loss explicitly adds discriminative constraints on a hypershpere manifold, which intrinsically matches the prior that human face lies on a manifold. However, Wang et al. \cite{Wang_2018_ECCV} showed that angular/cosine-margin-based loss can achieve better results on a clean dataset, but is vulnerable to noise and becomes worse than center loss and softmax in the high-noise region as shown in Fig. \ref{fig26}.

\begin{table}[htbp]
\small
\centering
\caption{Decision boundaries for class 1 under binary classification case, where $\hat{x}$ is the normalized feature. \cite{deng2019arcface}}
 \begin{tabular}{c|c}
  \hline
   Loss Functions                                            & Decision Boundaries \\ \hline \hline
   Softmax                                                   & $\left ( W_{1}-W_{2} \right )x+b_{1}-b_{2}=0$    \\ \hline
   L-Softmax \cite{liu2016large}                             & $\left \| x \right \|(\left \| W_{1} \right \|cos(m\theta _{1})- \left \| W_{2} \right \|cos(\theta _{2}))>0$\\ \hline
   A-Softmax \cite{liu2017sphereface}	                     & $\left \| x \right \|\left ( cosm\theta _{1}-cos\theta _{2} \right )=0 $   \\ \hline
   CosFace \cite{wang2018additive}                        & $\hat{x}\left ( cos\theta _{1}-m-cos\theta _{2} \right )=0$    \\ \hline
   ArcFace \cite{deng2019arcface}                            & $\hat{x}\left ( cos\left ( \theta _{1}+m \right )-cos\theta _{2} \right )=0$        \\ \hline

 \end{tabular}
 \label{tab6}
\end{table}

\subsubsection{ Softmax Loss and its Variations }

In 2017, in addition to reformulating softmax loss into an angular/cosine-margin-based loss as mentioned above, some works tries to normalize the features and weights in loss functions to improve the model performance, which can be written as follows:
\begin{equation}
\hat{W}=\frac{W}{\left \| W \right \|}, \hat{x}=\alpha\frac{x}{\left \| x \right \|}
\end{equation}
where $\alpha$ is a scaling parameter, $x$ is the learned feature vector, $W$ is weight of last fully connected layer. Scaling $x$ to a fixed radius $\alpha$ is important, as Wang  et al. \cite{wang2017normface} proved that normalizing both features and weights to 1 will make the softmax loss become trapped at a very high value on the training set. After that, the loss function, e.g. softmax, can be performed using the normalized features and weights. 

Some papers \cite{liu2017sphereface,liu2017deep} first normalized the weights only and then added angular/cosine margin into loss functions to make the learned features be discriminative. In contrast, some works, such as \cite{ranjan2017l2,hasnat2017deepvisage}, adopted feature normalization only to overcome the bias to the sample distribution of the softmax. Based on the observation of \cite{parde2016deep} that the L2-norm of features learned using the softmax loss is informative of the quality of the face, L2-softmax \cite{ranjan2017l2} enforced all the features to have the same L2-norm by feature normalization such that similar attention is given to good quality frontal faces and blurry faces with extreme pose. Rather than scaling $x$ to the parameter $\alpha$, Hasnat et al. \cite{hasnat2017deepvisage} normalized features with $\hat{x}=\frac{x-\mu}{\sqrt{\sigma^2}}$, where $\mu$ and $\sigma^2$ are the mean and variance. Ring loss \cite{Zheng_2018_CVPR} encouraged the norm of samples being value $R$ (a learned parameter) rather than explicit enforcing through a hard normalization operation. Moreover, normalizing both features and weights \cite{wang2017normface,liu2017rethinking,hasnat2017mises,wang2018additive,deng2019arcface} has become a common strategy. Wang et al. \cite{wang2017normface} explained the necessity of this normalization operation from both analytic and geometric perspectives. After normalizing features and weights, CoCo loss \cite{liu2017rethinking} optimized the cosine distance among data features, and Hasnat et al. \cite{hasnat2017mises} used the von Mises-Fisher (vMF) mixture model as the theoretical basis to develop a novel vMF mixture loss and its corresponding vMF deep features.


\subsection{Evolution of Network Architecture}

\subsubsection{ Backbone Network } \label{section}

\begin{figure}[htbp]
\centering
\includegraphics[width=9cm]{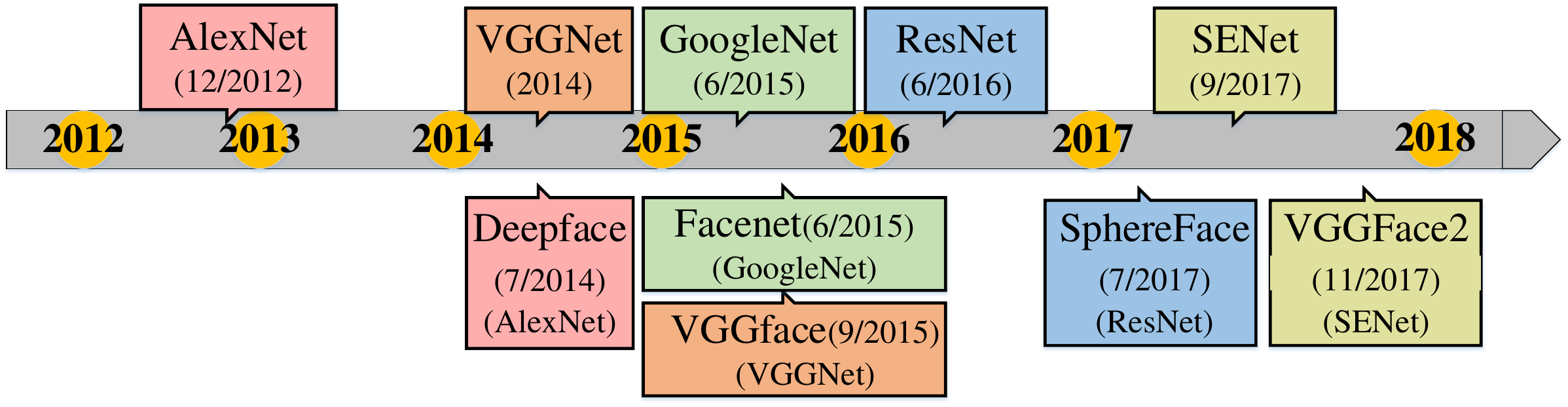}
\caption{The top row presents the typical network architectures in object classification, and the bottom row describes the well-known FR algorithms that use the typical architectures. We use the same color rectangles to represent the algorithms using the same architecture. It is easy to find that the architectures of deep FR have always followed those of deep object classification and evolved from AlexNet to SENet rapidly.}
\label{fig6}
\end{figure}

\textbf{Mainstream architectures.} The commonly used network architectures of deep FR have always followed those of deep object classification and evolved from AlexNet to SENet rapidly. We present the most influential architectures of deep object classification and deep face recognition in chronological order \footnote{The time we present is when the paper was published.} in Fig. \ref{fig6}. 

In 2012, AlexNet \cite{krizhevsky2012imagenet} was reported to achieve the SOTA recognition accuracy in the ImageNet large-scale visual recognition
competition (ILSVRC) 2012, exceeding the previous best results by a large margin. AlexNet consists of five convolutional layers and three fully connected layers, and it also integrates various techniques, such as rectified linear unit (ReLU), dropout, data augmentation, and so forth. ReLU was widely regarded as the most essential component for making deep learning possible. Then, in 2014, VGGNet \cite{simonyan2014very} proposed a standard network architecture that used very small $3\times3$ convolutional filters throughout and doubled the number of feature maps after the 2$\times$2 pooling. It increased the depth of the network to 16-19 weight layers, which further enhanced the flexibility to learn progressive nonlinear mappings by deep architectures. In 2015, the 22-layer GoogleNet \cite{szegedy2015going} introduced an ``inception module'' with the concatenation of hybrid feature maps, as well as two additional intermediate softmax supervised signals. It performs several convolutions with different receptive fields ($1\times1$, $3\times3$ and $5\times5$) in parallel, and concatenates all feature maps to merge the multi-resolution information. In 2016, ResNet \cite{he2016deep} proposed to make layers learn a residual mapping with reference to the layer inputs $\mathcal{F}(x):=\mathcal{H}(x)-x$ rather than directly learning a desired underlying mapping $\mathcal{H}(x)$ to ease the training of very deep networks (up to 152 layers). The original mapping is recast into $\mathcal{F}(x)+x$ and can be realized by ``shortcut connections''. As the champion of ILSVRC 2017, SENet \cite{hu2018squeeze} introduced a ``Squeeze-and-Excitation'' (SE) block, that adaptively recalibrates channel-wise feature responses by explicitly modelling interdependencies between channels. These blocks can be integrated with modern architectures, such as ResNet, and improves their representational power.

\begin{figure}[htbp]
\centering
\includegraphics[width=9cm]{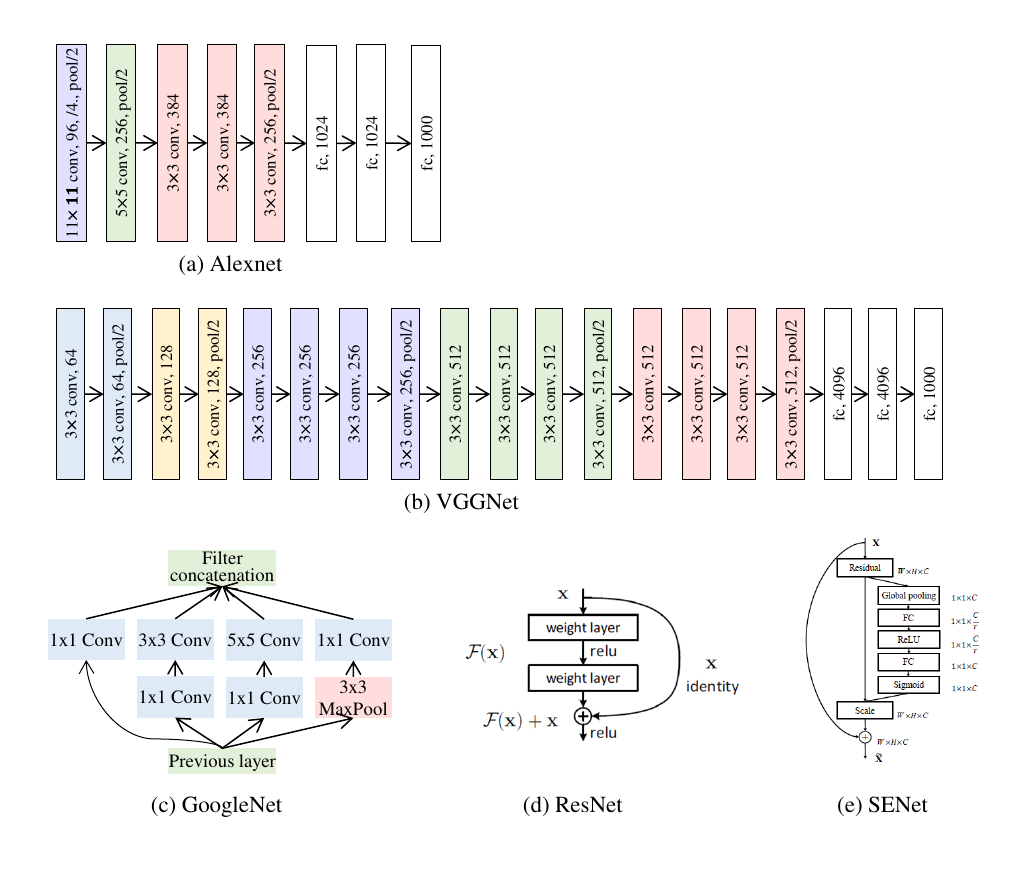}
\caption{The architecture of Alexnet \cite{krizhevsky2012imagenet}, VGGNet \cite{simonyan2014very}, GoogleNet \cite{szegedy2015going}, ResNet \cite{he2016deep}, SENet \cite{hu2018squeeze}.}
\label{fig7}
\end{figure}

With the evolved architectures and advanced training techniques, such as batch normalization (BN), the network becomes deeper and the training becomes more controllable. Following these architectures in object classification, the networks in deep FR are also developed step by step, and the performance of deep FR is continually improving. We present these mainstream architectures  of deep FR in Fig. \ref{fig7}. In 2014, DeepFace \cite{taigman2014deepface} was the first to use a nine-layer CNN with several locally connected layers. With 3D alignment for face processing, it reaches an accuracy of 97.35\% on LFW. In 2015, FaceNet \cite{schroff2015facenet} used a large private dataset to train a GoogleNet. It adopted a triplet loss function based on triplets of roughly aligned matching/nonmatching face patches generated by a novel online triplet mining method and achieved good performance of 99.63\%. In the same year, VGGface \cite{parkhi2015deep} designed a procedure to collect a large-scale dataset from the Internet. It trained the VGGNet on this dataset and then fine-tuned the networks via a triplet loss function similar to FaceNet. VGGface obtains an accuracy of 98.95\%. In 2017, SphereFace \cite{liu2017sphereface} used a 64-layer ResNet architecture and proposed the angular softmax (A-Softmax) loss to learn discriminative face features with angular margin. It boosts the achieves to 99.42\% on LFW. In the end of 2017, a new large-scale face dataset, namely VGGface2 \cite{cao2018vggface2}, was introduced, which consists of large variations in pose, age, illumination, ethnicity and profession. Cao et al. first trained a SENet with MS-celeb-1M dataset \cite{guo2016ms} and then fine-tuned the model with VGGface2 \cite{cao2018vggface2}, and achieved the SOTA performance on the IJB-A \cite{klare2015pushing} and IJB-B \cite{Whitelam2017IARPA}.

\textbf{Light-weight networks.} Using deeper neural network with hundreds of layers and millions of parameters to achieve higher accuracy comes at cost. Powerful GPUs with larger memory size are needed, which makes the applications on many mobiles and embedded devices impractical. To address this problem, light-weight networks are proposed. Light CNN \cite{wu2018light,wu2015lightened} proposed a max-feature-map (MFM) activation function that introduces the concept of maxout in the fully connected layer to CNN. The MFM obtains a compact representation and reduces the computational cost. Sun et al. \cite{sun2016sparsifying} proposed to sparsify deep networks iteratively from the previously learned denser models based on a weight selection criterion. MobiFace \cite{duong2018mobiface} adopted fast downsampling and bottleneck residual block with the expansion layers and achieved
high performance with 99.7\% on LFW database. 
Although some other light-weight CNNs, such as SqueezeNet, MobileNet, ShuffleNet and Xception \cite{iandola2016squeezenet,howard2017mobilenets,chollet2017xception,zhang2018shufflenet}, are still not widely used in FR, they deserve more attention.

\textbf{Adaptive-architecture networks.} Considering that designing architectures manually by human experts are time-consuming and error-prone processes, there is growing interest in adaptive-architecture networks which can find well-performing architectures, e.g.  the type of operation every layer executes (pooling, convolution, etc) and hyper-parameters associated with the operation (number of filters, kernel size and strides for a convolutional layer, etc), according to the specific requirements of training and testing data. Currently, neural architecture search (NAS) \cite{zoph2016neural} is one of the promising methodologies, which has outperformed manually designed architectures on some tasks such as image classification \cite{real2019aging} or semantic segmentation \cite{chen2018searching}. Zhu et al. \cite{zhu2020new} integrated NAS technology into face recognition. They used reinforcement learning \cite{mnih2015human} algorithm (policy gradient) to guide the controller network to train the optimal child architecture. Besides NAS, there are some other explorations to learn optimal architectures adaptively. For example, conditional convolutional neural network (c-CNN) \cite{xiong2015conditional} dynamically activated sets of kernels according to modalities of samples; Han et al. \cite{Han_2018_ECCV} proposed a novel contrastive convolution consisted of a trunk CNN and a kernel generator, which is beneficial owing to its dynamistic generation of contrastive kernels based on the pair of faces being compared.

\begin{figure}[htbp]
\centering
\includegraphics[width=9cm]{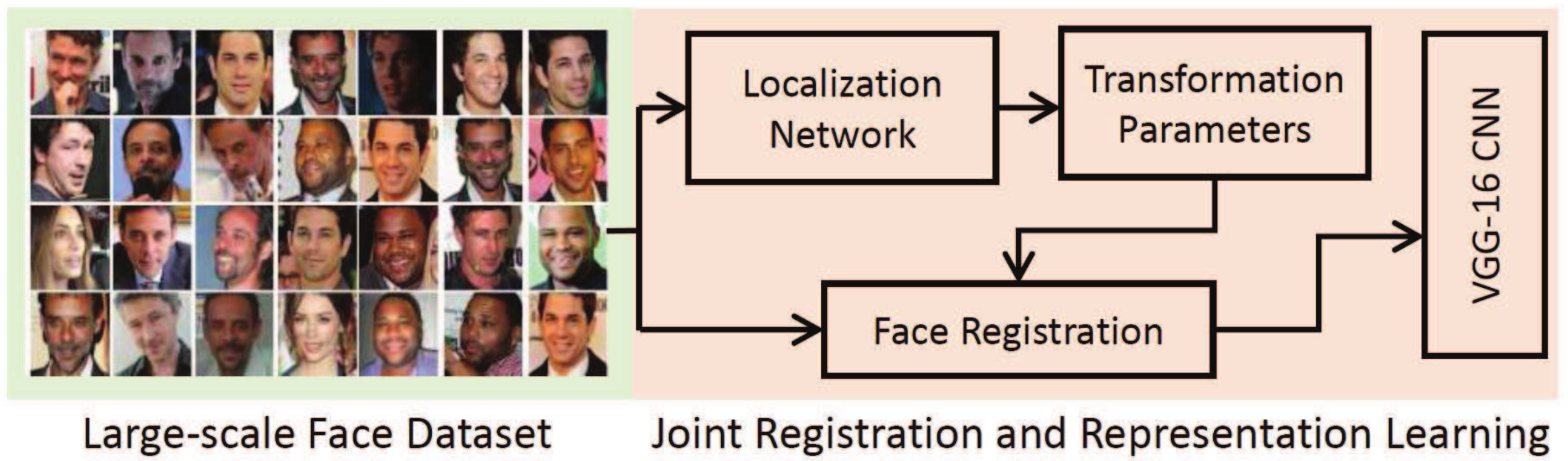}
\caption{ Joint face registration and representation learning. \cite{hayat2017joint}}
\label{fig10}
\end{figure}

\textbf{Joint alignment-recognition networks.} Recently, an end-to-end system \cite{hayat2017joint,wu2017recursive,zhong2017toward,chen2015end} was proposed to jointly train FR with several modules (face detection, alignment, and so forth) together. Compared to the existing methods in which each module is generally optimized separately according to different objectives, this end-to-end system optimizes each module according to the recognition objective, leading to more adequate and robust inputs for the recognition model. For example, inspired by spatial transformer \cite{jaderberg2015spatial}, Hayat et al. \cite{hayat2017joint} proposed a CNN-based data-driven approach that learns to simultaneously register and represent faces (Fig. \ref{fig10}), while Wu et al. \cite{wu2017recursive} designed a novel recursive spatial transformer (ReST) module for CNN allowing face alignment and recognition to be jointly optimized.


\subsubsection{ Assembled Networks } \label{Multiple Network}

\textbf{Multi-input networks.} In ``one-to-many augmentation'', multiple images with variety are generated from one image in order to augment training data. Taken these multiple images as input, multiple networks are also assembled together to extract and combine features of different type of inputs, which can outperform an individual network. In \cite{liu2015targeting,zhou2015naive,ding2015robust,sun2013hybrid,sun2014deep1,sun2014deep,sun2015deeply}, assembled networks are built after different face patches are cropped, and then different types of patches are fed into different sub-networks for representation extraction. By combining the results of sub-networks, the performance can be improved. Other papers \cite{masi2016pose,kan2016multi,wang2014deeply} used assembled networks to recognize images with different poses. For example, Masi et al. \cite{masi2016pose} adjusted the pose to frontal ($0^{\circ}$), half-profile ($40^{\circ}$) and full-profile views ($75^{\circ}$) and then addressed pose variation by assembled pose networks. A multi-view deep network (MvDN) \cite{kan2016multi} consists of view-specific subnetworks and common subnetworks; the former removes view-specific variations, and the latter obtains common representations. 

\begin{figure}[htbp]
\centering
\includegraphics[width=9cm]{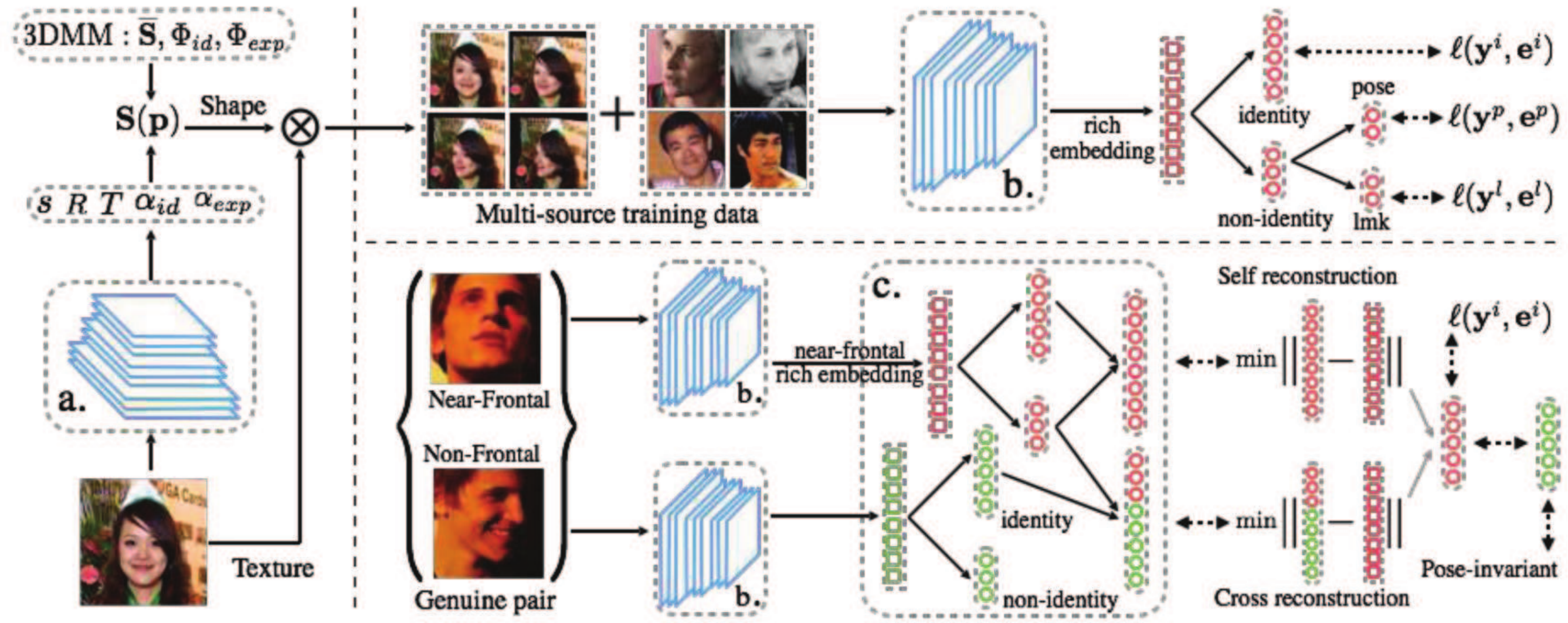}
\caption{Reconstruction-based disentanglement for pose-invariant FR. \cite{peng2017reconstruction}}
\label{fig8}
\end{figure}

\textbf{Multi-task networks.} FR is intertwined with various factors, such as pose, illumination, and age. To solve this problem, multitask learning is introduced to transfer knowledge from other relevant tasks and to disentangle nuisance factors. In multi-task networks, identity classification is the main task and the side tasks are pose, illumination, and expression estimations, among others. The lower layers are shared among all the tasks, and the higher layers are disentangled into different sub-networks to generate the task-specific outputs. In \cite{ranjan2017all}, the task-specific sub-networks are branched out to learn face detection, face alignment, pose estimation, gender recognition, smile detection, age estimation and FR. Yin et al. \cite{yin2017multi} proposed to automatically assign the dynamic loss weights for each side task. Peng et al. \cite{peng2017reconstruction} used a feature reconstruction metric learning to disentangle a CNN into sub-networks for jointly learning the identity and non-identity features as shown in Fig. \ref{fig8}.

\subsection{Face Matching by deep features}

During testing, the cosine distance and L2 distance are generally employed to measure the similarity between the deep features $x_{1}$ and $x_{2}$; then, threshold comparison and the nearest neighbor (NN) classifier are used to make decision for verification and identification. In addition to these common methods, there are some other explorations.

\subsubsection{Face verification}

Metric learning, which aims to find a new metric to make two classes more separable, can also be used for face matching based on extracted deep features. The JB \cite{chen2012bayesian} model is a well-known metric learning method \cite{sun2015deeply,sun2014deep,sun2015deepid3,sun2014deep1,yi2014learning}, and Hu et al. \cite{hu2015face} proved that it can improve the performance greatly. In the JB model, a face feature $x$ is modeled as $x=\mu+\varepsilon$, where $\mu$ and $\varepsilon$ are identity and intra-personal variations, respectively. The similarity score $r(x_{1},x_{2})$ can be represented as follows:
\begin{equation}
r(x_{1},x_{2})=log\frac{P\left ( x_{1},x_{2}|H_{I} \right )}{P\left ( x_{1},x_{2}|H_{E} \right )}
\end{equation}
where $P(x_{1},x_{2}|H_{I})$ is the probability that two faces belong to the same identity and $P(x_{1},x_{2}|H_{E})$ is the probability that two faces belong to different identities.

\subsubsection{Face identification}

After cosine distance was computed, Cheng et al. \cite{cheng2017know} proposed a heuristic voting strategy at the similarity score level to combine the results of multiple CNN models and won first place in Challenge 2 of MS-celeb-1M 2017. Yang et al. \cite{Yang2016Joint} extracted the local adaptive convolution features from the local regions of the face image and used the extended SRC for FR with a single sample per person. Guo et al. \cite{Guo2017Face} combined deep features and the SVM classifier to perform recognition. Wang et al. \cite{wang2016face} first used product quantization (PQ) \cite{Jegou2011Product} to directly retrieve the top-k most similar faces and re-ranked these faces by combining similarities from deep features and the COTS matcher \cite{Grother2014Face}. In addition, Softmax can be also used in face matching when the identities of training set and test set overlap. For example, in Challenge 2 of MS-celeb-1M, Ding et al. \cite{Ding2018Generative} trained a 21,000-class softmax classifier to directly recognize faces of one-shot classes and normal classes after augmenting feature by a conditional GAN; Guo et al. \cite{guo2017one} trained the softmax classifier combined with underrepresented-classes promotion (UP) loss term to enhance the performance on one-shot classes.

When the distributions of training data and testing data are the same, the face matching methods mentioned above are effective. However, there is always a distribution change or domain shift between two data domains that can degrade the performance on test data. Transfer learning \cite{pan2010survey,wang2018deep} has recently been introduced into deep FR to address the problem of domain shift. It learns transferable features using a labeled source domain (training data) and an unlabeled target domain (testing data) such that domain discrepancy is reduced and models trained on source domain will also perform well on target domain. Sometimes, this technology is applied to face matching. For example, Crosswhite et al. \cite{crosswhite2017template} and Xiong et al. \cite{xiong2017good} adopted template adaptation to the set of media in a template by combining CNN features with template-specific linear SVMs. But most of the time, it is not enough to do transfer learning only at face matching stage. Transfer learning should be embedded in deep models to learn more transferable representations. Kan et al. \cite{Kan2015Bi} proposed a bi-shifting autoencoder network (BAE) for domain adaptation across view angle, ethnicity, and imaging sensor; while Luo et al. \cite{Luo2018Adaptation} utilized the multi-kernels maximum mean discrepancy (MMD) to reduce domain discrepancies. Sohn et al. \cite{Sohn2017Unsupervised} used adversarial learning \cite{tzeng2017adversarial} to transfer knowledge from still image FR to video FR. Moreover, fine-tuning the CNN parameters from a prelearned model using a target training dataset is a particular type of transfer learning, and is commonly employed by numerous methods \cite{Abdalmageed2016Face,wang2017face,chen2016unconstrained}.

\section{Face Processing for Training and Recognition} \label{Pre-processing}

\begin{figure}[htbp]
\centering
\includegraphics[width=9cm]{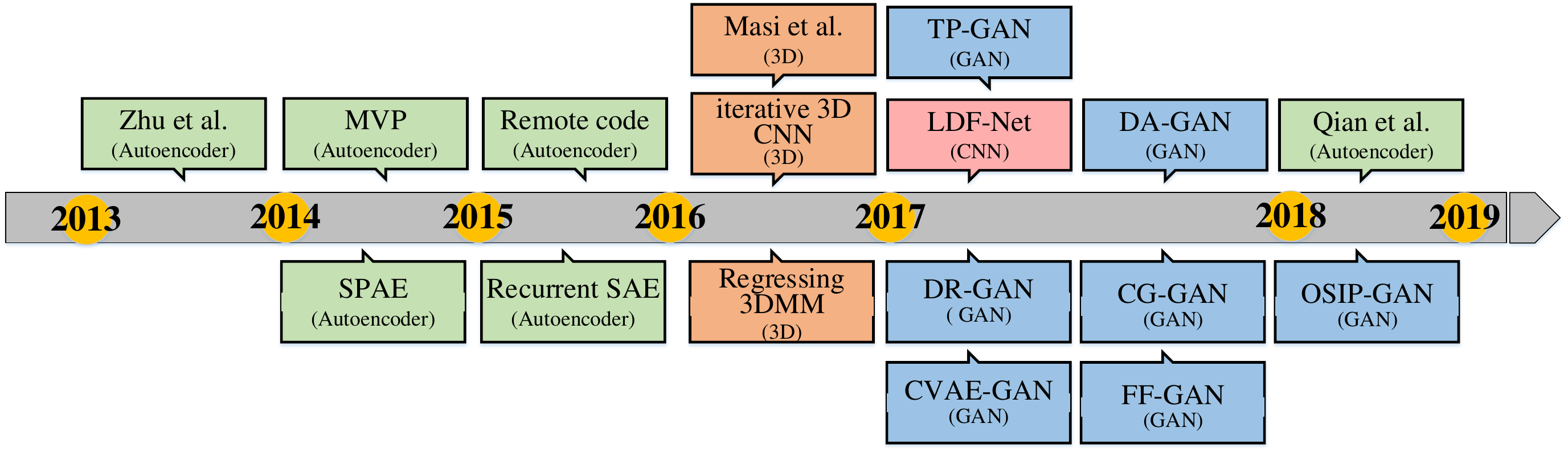}
\caption{The development of deep face processing methods. Red, green, orange and blue rectangles represent CNN model, autoencoder model, 3D model and GAN model, respectively.}
\label{fig24}
\end{figure}

We present the development of face processing methods in chronological order in Fig. \ref{fig24}. As we can see from the figure, most papers attempted to perform face processing by autoencoder model in 2014 and 2015; while 3D model played an important role in 2016. GAN \cite{goodfellow2014generative} has drawn substantial attention from the deep learning and computer vision community since it was first proposed by Goodfellow et al. It can be used in different fields and was also introduced into face processing in 2017. GAN can be used to perform ``one-to-many augmentation'' and ``many-to-one normalization'', and it broke the limit that face synthesis should be done under supervised way. Although GAN has not been widely used in face processing for training and recognition, it has great latent capacity for preprocessing, for example, Dual-Agent GANs (DA-GAN) \cite{zhao2017dual} won the 1st places on verification and identification tracks in the NIST IJB-A 2017 FR competitions.

\subsection{One-to-Many Augmentation}

Collecting a large database is extremely expensive and time consuming. The methods of ``one-to-many augmentation'' can mitigate the challenges of data collection, and they can be used to augment not only training data but also the gallery of test data. we categorized them into four classes: data augmentation, 3D model, autoencoder model and GAN model.

\textbf{Data augmentation}. Common data augmentation methods consist of photometric transformations \cite{simonyan2014very,krizhevsky2012imagenet} and geometric transformations, such as oversampling (multiple patches obtained by cropping at different scales) \cite{krizhevsky2012imagenet}, mirroring \cite{yang2015mirror}, and rotating \cite{xie2015holistically} the images. Recently, data augmentation has been widely used in deep FR algorithms \cite{liu2015targeting,zhou2015naive,ding2015robust,sun2015deeply,sun2014deep,sun2015deepid3,sun2016sparsifying,wang2016face}. for example, Sun et al. \cite{sun2014deep} cropped 400 face patches varying in positions, scales, and color channels and mirrored the images. Liu et al. \cite{liu2015targeting} generated seven overlapped image patches centered at different landmarks on the face region and trained them with seven CNNs with the same structure.

\begin{figure}[htbp]
\centering
\includegraphics[width=9cm]{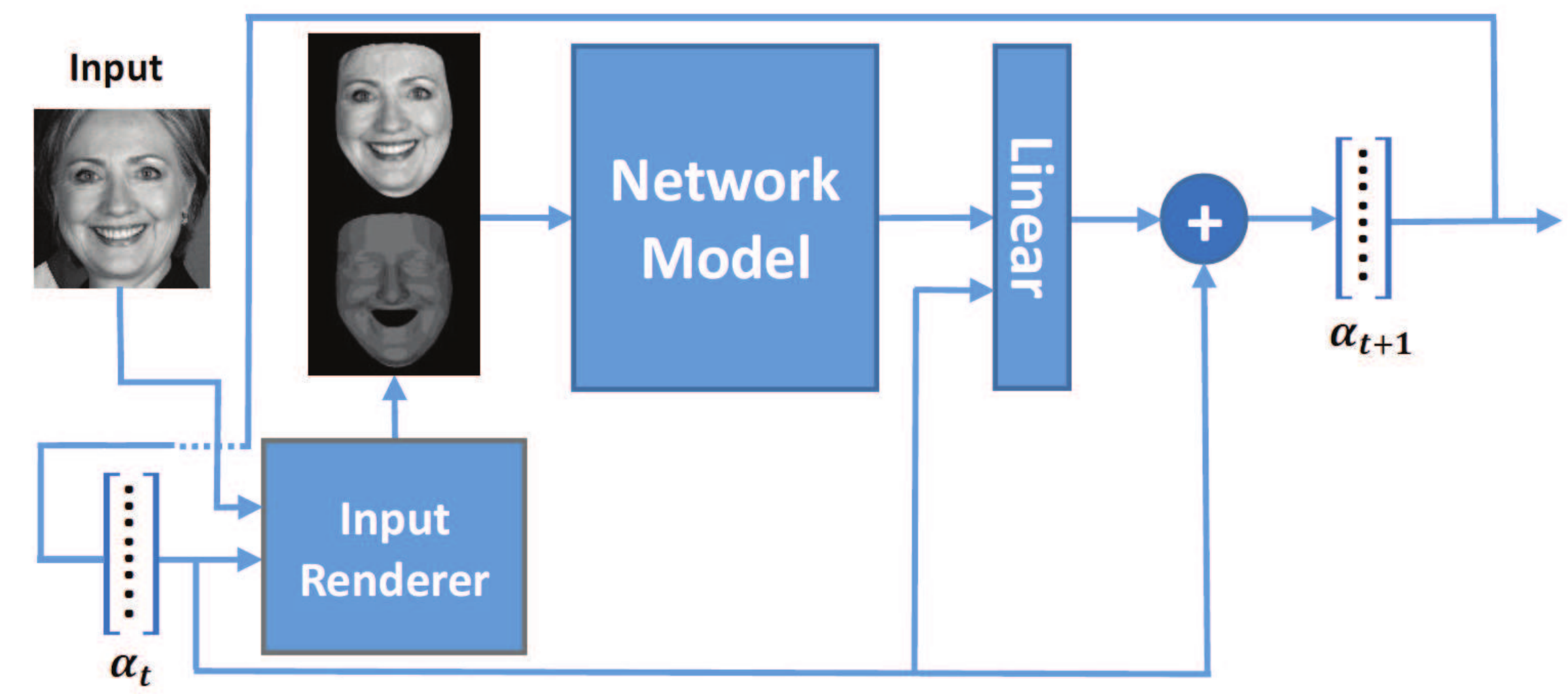}
\caption{Iterative CNN network for reconstructing a 3D face. \cite{richardson20163d}}
\label{fig4}
\end{figure}

\textbf{3D model}. 3D face reconstruction is also a way to enrich the diversity of training data. They utilize 3D structure information to model the transformation between poses. 3D models first use 3D face data to obtain morphable displacement fields and then apply them to obtain 2D face data in different pose angles. There is a large number of papers about this domain, but we only focus on the 3D face reconstruction using deep methods or used for deep FR. In \cite{masi2016we}, Masi et al. generated face images with new intra-class facial appearance variations, including pose, shape and expression, and then trained a 19-layer VGGNet with both real and augmented data. Masi et al. \cite{masi2017rapid} used generic 3D faces and rendered fixed views to reduce much of the computational effort. Richardson et al. \cite{richardson20163d} employed an iterative 3D CNN by using a secondary input channel to represent the previous network's output as an image for reconstructing a 3D face as shown in Fig. \ref{fig4}.  Dou et al. \cite{dou2017end} used a multi-task CNN to divide 3D face reconstruction into neutral 3D reconstruction and expressive 3D reconstruction. Tran et al. \cite{tuan2017regressing} directly regressed 3D morphable face model (3DMM) \cite{blanz2003face} parameters from an input photo by a very deep CNN architecture. An et al. \cite{An2018Transfer} synthesized face images with various poses and expressions using the 3DMM method, then reduced the gap between synthesized data and real data with the help of MMD.

\begin{figure}[htbp]
\centering
\includegraphics[width=9cm]{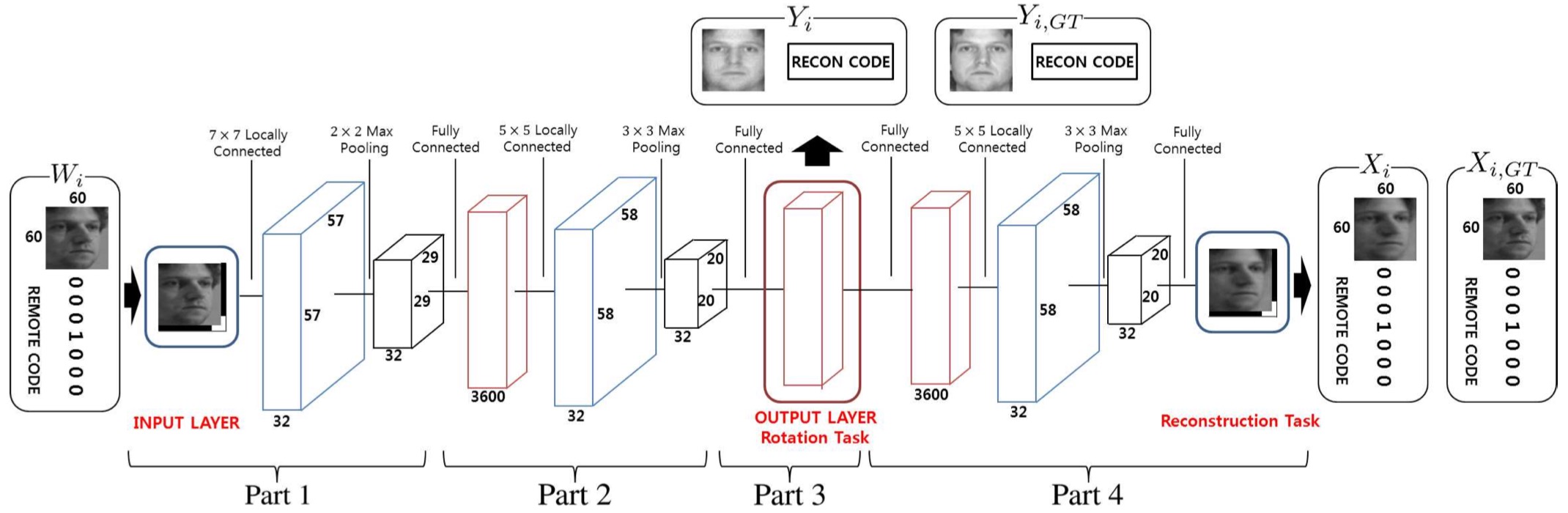}
\caption{Autoencoder model of ``one-to-many augmentation'' proposed by \cite{yim2015rotating}. The first part extracts feature from an input image, then the second and third part generate a target image with the same identity viewed at the target pose. The forth part is an auxiliary task which reconstructs the original input image back from the generated image to guarantee that the generated image is identity-preserving.}
\label{autoencoder}
\end{figure}

\textbf{Autoencoder model}. Rather than reconstructing 3D models from a 2D image and projecting it back into 2D images of different poses, autoencoder models can generate 2D target images directly. Taken a face image and a pose code encoding a target pose as input, an encoder first learns pose-invariant face representation, and then a decoder generates a face image with the same identity viewed at the target pose by using the pose-invariant representation and the pose code. For example, given the target pose codes, multi-view perceptron (MVP) \cite{zhu2014multi} trained some deterministic hidden neurons to learn pose-invariant face representations, and simultaneously trained some random hidden neurons to capture pose features, then a decoder generated the target images by combining pose-invariant representations with pose features. As shown in Fig. \ref{autoencoder}, Yim et al. \cite{yim2015rotating} and Qian et al. \cite{Qian2018Task} introduced an auxiliary CNN to generate better images viewed at the target poses. First, an autoencoder generated the desired pose image, then the auxiliary CNN reconstructed the original input image back from the generated target image, which guarantees that the generated image is identity-preserving. In \cite{yang2015weakly}, two groups of units are embedded between encoder and decoder. The identity units remain unchanged and the rotation of images is achieved by taking actions to pose units at each time step.

\textbf{GAN model}. In GAN models, a generator aims to fool a discriminator through generating images that resemble the real images, while the discriminator aims to discriminate the generated samples from the real ones. By this minimax game between generator and discriminator, GAN can successfully generate photo-realistic images with different poses. After using a 3D model to generate profile face images, DA-GAN \cite{zhao2017dual} refined the images by a GAN, which combines prior knowledge of the data distribution and knowledge of faces (pose and identity perception loss). CVAE-GAN \cite{bao2017cvae} combined a variational auto-encoder with a GAN for augmenting data, and took advantages of both statistic and pairwise feature matching to make the training process converge faster and more stably. In addition to synthesizing diverse faces from noise, some papers also explore to disentangle the identity and variation, and synthesize new faces by exchanging identity and variation from different people. In CG-GAN \cite{Chai2018Cross}, a generator directly resolves each representation of input image into a variation code and an identity code and regroups these codes for cross-generating, simultaneously, a discriminator ensures the reality of generated images. Bao et al. \cite{bao2018towards} extracted identity representation of one input image and attribute representation of any other input face image, then synthesized new faces by recombining these representations. This work shows superior performance in generating realistic and identity preserving face images, even for identities outside the training dataset. Unlike previous methods that treat classifier as a spectator, FaceID-GAN \cite{Shen2018faceid} proposed a three-player GAN where the classifier cooperates together with the discriminator to compete with the generator from two different aspects, i.e. facial identity and image quality respectively.

\subsection{ Many-to-One Normalization}

In contrast to ``one-to-many augmentation'', the methods of ``many-to-one normalization'' produce frontal faces and reduce appearance variability of test data to make faces align and compare easily. It can be categorized as autoencoder model, CNN model and GAN model.

\textbf{Autoencoder model}. Autoencoder can also be applied to ``many-to-one normalization''. Different from the autoencoder model in ``one-to-many augmentation'' which generates the desired pose images with the help of pose codes, autoencoder model here learns pose-invariant face representation by an encoder and directly normalizes faces by a decoder without pose codes. Zhu et al. \cite{zhu2013deep,zhu2014recover} selected canonical-view images according to the face images' symmetry and sharpness and then adopted an autoencoder to recover the frontal view images by minimizing the reconstruction loss error. The proposed stacked progressive autoencoders (SPAE) \cite{kan2014stacked} progressively map the nonfrontal face to the frontal face through a stack of several autoencoders. Each shallow autoencoders of SPAE is designed to convert the input face images at large poses to a virtual view at a smaller pose, so the pose variations are narrowed down gradually layer by layer along the pose manifold. Zhang et al. \cite{zhang2013random} built a sparse many-to-one encoder to enhance the discriminant of the pose free feature by using multiple random faces as the target values for multiple encoders.

\begin{figure}[htbp]
\centering
\includegraphics[width=9cm]{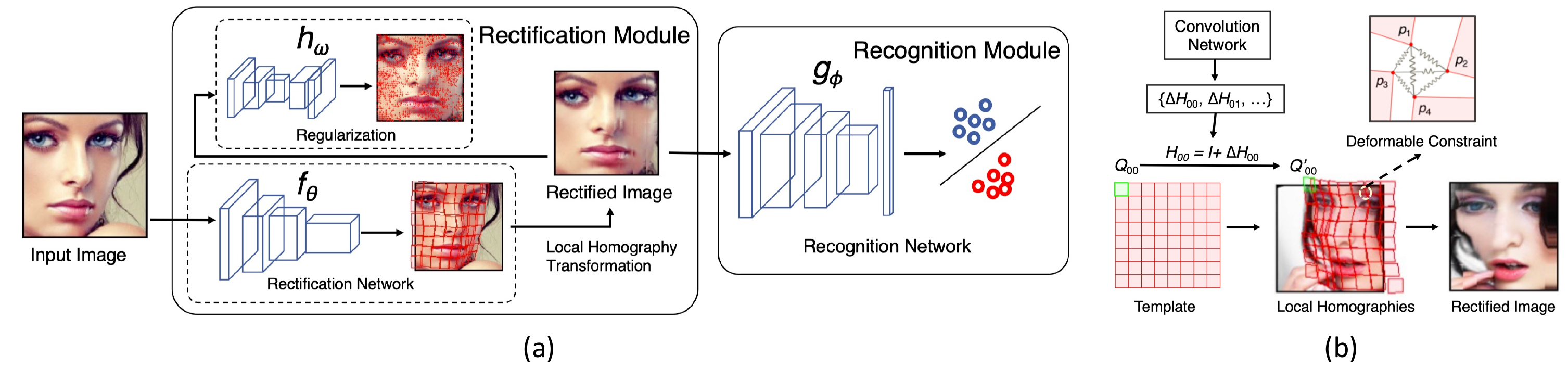}
\caption{(a) System overview and (b) local homography transformation of GridFace \cite{Zhou_2018_ECCV}. The rectification network normalizes the images by warping pixels from the original image to the canonical one according to the computed homography matrix.}
\label{cnn_many_one}
\end{figure}

\textbf{CNN model}. CNN models usually directly learn the 2D mappings between non-frontal face images and frontal images, and utilize these mapping to normalize images in pixel space. The pixels in normalized images are either directly the pixels or the combinations of the pixels in non-frontal images. In LDF-Net \cite{hu2017ldf}, the displacement field network learns the shifting relationship of two pixels, and the translation layer transforms the input non-frontal face image into a frontal one with this displacement field.  In GridFace \cite{Zhou_2018_ECCV} shown in Fig. \ref{cnn_many_one}, first, the rectification network normalizes the images by warping pixels from the original image to the canonical one according to the computed homography matrix, then the normalized output is regularized by an implicit canonical view face prior, finally, with the normalized faces as input, the recognition network learns discriminative face representation via metric learning.

\begin{figure}[htbp]
\centering
\includegraphics[width=9cm]{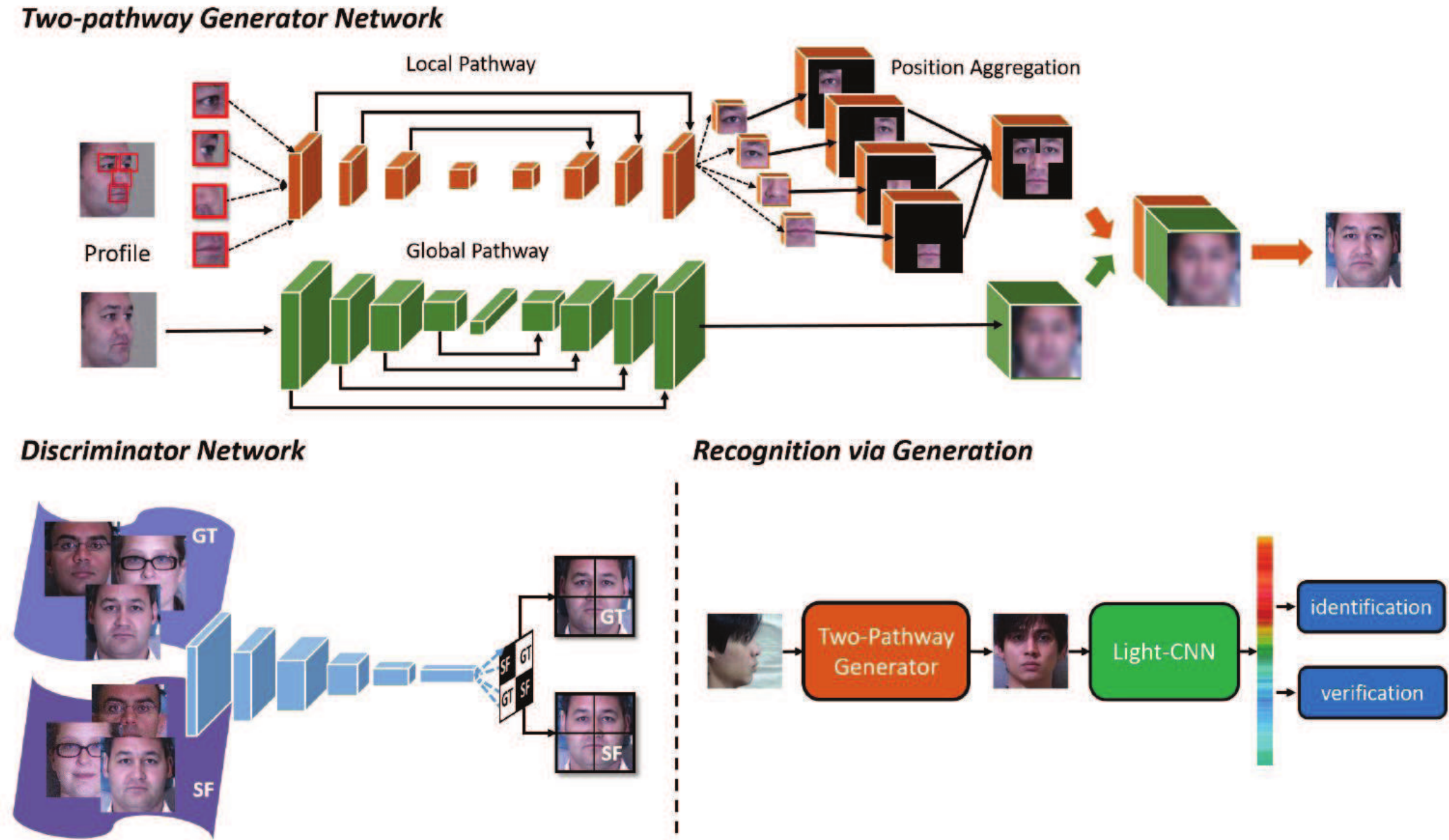}
\caption{General framework of TP-GAN \cite{huang2017beyond}. The generator contains two pathways with each processing global or local transformations. The discriminator distinguishes between synthesized frontal views and ground-truth frontal views.}
\label{fig5}
\end{figure}

\textbf{GAN model}. Huang et al. \cite{huang2017beyond} proposed a two-pathway generative adversarial network (TP-GAN) that contains four landmark-located patch networks and a global encoder-decoder network. Through combining adversarial loss, symmetry loss and identity-preserving loss, TP-GAN generates a frontal view and simultaneously preserves global structures and local details as shown in Fig. \ref{fig5}. In a disentangled representation learning generative adversarial network (DR-GAN) \cite{tran2017disentangled}, the generator serves as a face rotator, in which an encoder produces an identity representation, and a decoder synthesizes a face at the specified pose using this representation and a pose code. And the discriminator  is trained to not only distinguish real vs. synthetic images, but also predict the identity and pose of a face. Yin et al. \cite{yin2017towards} incorporated 3DMM into the GAN structure to provide shape and appearance priors to guide the generator to frontalization.

\section{Face databases and Evaluation Protocols}

\begin{figure*}
\centering
\includegraphics[width=\textwidth]{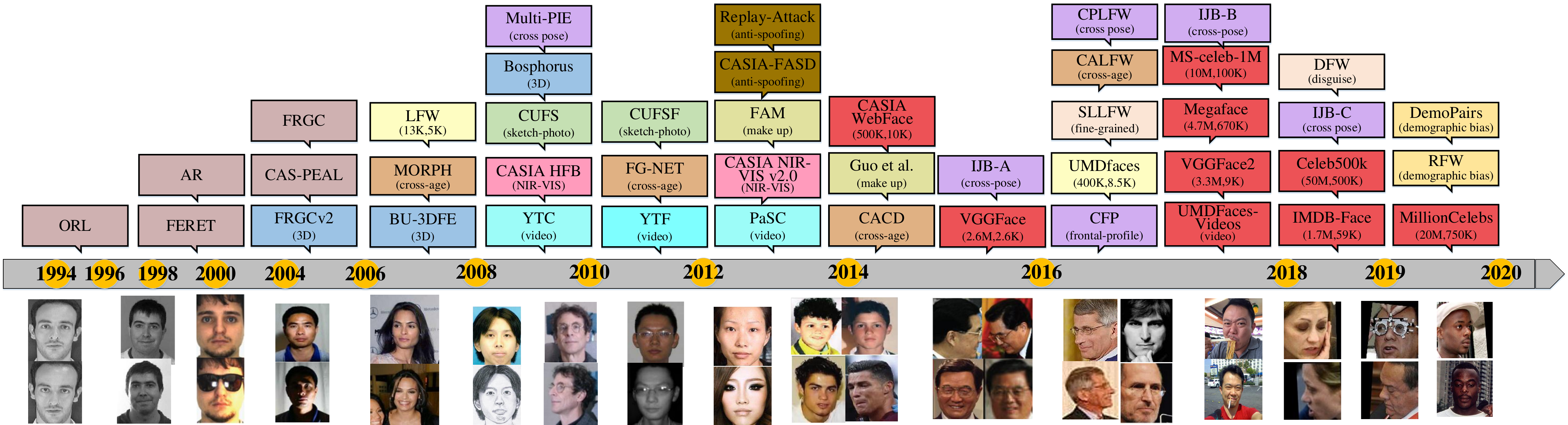}
\caption{The evolution of FR datasets. Before 2007, early works in FR focused on controlled and small-scale datasets. In 2007, LFW \cite{huang2007labeled} dataset was introduced which marks the beginning of FR under unconstrained conditions. Since then, more testing databases designed for different tasks and scenes are constructed. And in 2014, CASIA-Webface \cite{yi2014learning} provided the first widely-used public training dataset, large-scale training datasets begun to be hot topic. Red rectangles represent training datasets, and other color rectangles represent different testing datasets. }
\label{fig22}
\end{figure*}

In the past three decades, many face databases have been constructed with a clear tendency from small-scale to large-scale, from single-source to diverse-sources, and from lab-controlled to real-world unconstrained condition, as shown in Fig. \ref{fig22}. As the performance of some simple databases become saturated, e.g. LFW  \cite{huang2007labeled}, more and more complex databases were continually developed to facilitate the FR research. It can be said without exaggeration that the development process of the face databases largely leads the direction of FR research. In this section, we review the development of major training and testing academic databases for the deep FR.

\subsection{Large-scale training data sets} \label{dataset_ref}

The prerequisite of effective deep FR is a sufficiently large training dataset. Zhou et al. \cite{zhou2015naive} suggested that large amounts of data with deep learning improve the performance of FR. The results of Megaface Challenge also revealed that premier deep FR methods were typically trained on data larger than 0.5M images and 20K people. The early works of deep FR were usually trained on private training datasets. Facebook's Deepface \cite{taigman2014deepface} model was trained on 4M images of 4K people; Google's FaceNet \cite{schroff2015facenet} was trained on 200M images of 3M people; DeepID serial models \cite{sun2014deep1,sun2015deeply,sun2014deep,sun2015deepid3} were trained on 0.2M images of 10K people. Although they reported ground-breaking performance at this stage, researchers cannot accurately reproduce or compare their models without public training datasets.

\begin{table*}[htbp]
\scriptsize
\centering
\begin{threeparttable}
\caption{The commonly used FR datasets for training}
\setlength{\tabcolsep}{0mm}{
 \begin{tabular}{c|c|c|c|c|c}
  \hline
   Datasets                       & \tabincell{l}{Publish\\ Time}   & \#photos        & \#subjects  &\tabincell{l}{\# of photos \\per subject \tnote{1}} & Key Features \\ \hline \hline
   \tabincell{c}{MS-Celeb-1M\\(Challenge 1)\cite{guo2016ms}} & 2016  & \tabincell{c}{10M\\3.8M(clean)}	           & \tabincell{c}{100,000\\85K(clean)}	 & 100  & \tabincell{c}{ breadth; central part of long tail;\\ celebrity; knowledge base} \\ \hline
   \tabincell{c}{MS-Celeb-1M\\(Challenge 2)\cite{guo2016ms}} & 2016  & \tabincell{c}{1.5M(base set)\\1K(novel set)}	           & \tabincell{c}{20K(base set)\\1K(novel set)}	 & 1/-/100  & \tabincell{c}{low-shot learning; tailed data; \\celebrity}  \\ \hline
   \tabincell{c}{MS-Celeb-1M\\(Challenge 3) \cite{MS3}} & 2018  & \tabincell{c}{4M(MSv1c)\\2.8M(Asian-Celeb)}	   & \tabincell{c}{80K(MSv1c)\\100K(Asian-Celeb)}	 & -  &  \tabincell{c}{breadth;central part of long tail;\\celebrity }\\ \hline
   MegaFace	\cite{kemelmacher2016megaface,nech2017level}     & 2016  & 4.7M	           & 672,057	 & 3/7/2469    & \tabincell{c}{ breadth; the whole long \\tail;commonalty}\\ \hline
   VGGFace2 \cite{cao2018vggface2}                           & 2017  & 3.31M           & 9,131       & 87/362.6/843       & \tabincell{c}{depth; head part of long tail; cross \\ pose, age and ethnicity; celebrity} \\ \hline
   CASIA WebFace \cite{yi2014learning}	                     & 2014  & 494,414	       & 10,575	     & 2/46.8/804    &celebrity\\ \hline
   MillionCelebs \cite{zhang2020global}                         &2020   &18.8M                  & 636K            & 29.5             & celebrity\\ \hline
   IMDB-Face \cite{Wang_2018_ECCV}                        &2018    & 1.7M            & 59K            & 28.8                  & celebrity\\ \hline
   UMDFaces-Videos \cite{bansal2017s}                        & 2017  & 22,075          & 3,107       & --     & video \\ \hline
   VGGFace \cite{parkhi2015deep}	                         & 2015  & 2.6M	           & 2,622       & 1,000    & \tabincell{c}{depth; celebrity; annotation with \\bounding boxesand coarse pose}\\ \hline
   CelebFaces$+$ \cite{sun2014deep}                          & 2014  & 202,599	       & 10,177      & 19.9      &private\\ \hline
   Google \cite{schroff2015facenet}	                         & 2015  & $>$500M	       & $>$10M	     & 50	    & private \\ \hline
   Facebook \cite{taigman2014deepface}                       & 2014  & 4.4M	           & 4K	         & 800/1100/1200	   & private\\ \hline

 \end{tabular}}
 \label{tab0}
 \begin{tablenotes}
  \item[1] The min/average/max numbers of photos or frames per subject
\end{tablenotes}
 \end{threeparttable}
\end{table*}

\begin{figure}[htbp]
\centering
\includegraphics[width=9cm]{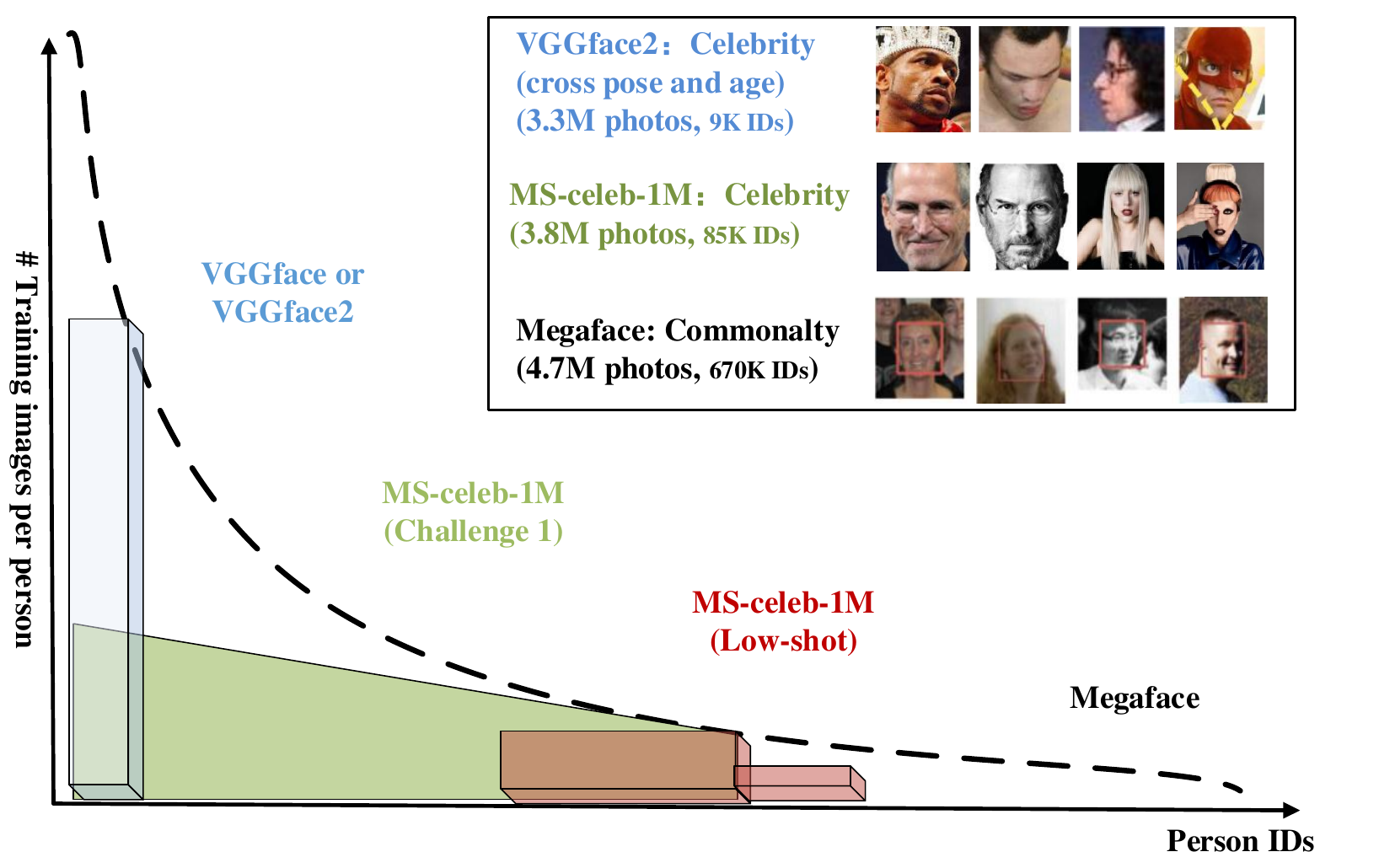}
\caption{ The distribution of three new large-scale databases suitable for training deep models. They have larger scale than the widely-used CAISA-Web database. The vertical axis displays number of images per person, and the horizontal axis shows person IDs.}
\label{fig18}
\end{figure}

To address this issue, CASIA-Webface \cite{yi2014learning} provided the first widely-used public training dataset for the deep model training purpose, which consists of 0.5M images of 10K celebrities collected from the web. Given its moderate size and easy usage, it has become a great resource for fair comparisons for academic deep models. However, its relatively small data and ID size may not be sufficient to reflect the power of many advanced deep learning methods. Currently, there have been more databases providing public available large-scale training dataset (Table \ref{tab0}), especially three databases with over 1M images, namely MS-Celeb-1M \cite{guo2016ms}, VGGface2 \cite{cao2018vggface2}, and Megaface	 \cite{kemelmacher2016megaface,nech2017level}, and we summary some interesting findings about these training sets, as shown in Fig. \ref{fig18}.

\textbf{Depth v.s. breadth.} These large training sets are expanded from depth or breadth. VGGface2 provides a large-scale training dataset of depth, which have limited number of subjects but many images for each subjects. The depth of dataset enforces the trained model to address a wide range intra-class variations, such as lighting, age, and pose. In contrast, MS-Celeb-1M and Mageface (Challenge 2) offers large-scale training datasets of breadth, which contains many subject but limited images for each subjects. The breadth of dataset ensures the trained model to cover the sufficiently variable appearance of various people. Cao et al. \cite{cao2018vggface2} conducted a systematic studies on model training using VGGface2 and MS-Celeb-1M, and found an optimal model by first training on MS-Celeb-1M (breadth) and then fine-tuning on VGGface2 (depth).

\textbf{Long tail distribution.} The utilization of long tail distribution is different among datasets. For example, in Challenge 2 of MS-Celeb-1M, the novel set specially uses the tailed data to study low-shot learning; 
central part of the long tail distribution is used by the Challenge 1 of MS-Celeb-1M and images' number is approximately limited to 100 for each celebrity; VGGface and VGGface2 only use the head part to construct deep databases; Megaface utilizes the whole distribution to contain as many images as possible, the minimal number of images is 3 per person and the maximum is 2469.

\textbf{Data engineering.} Several popular benchmarks, such as LFW unrestricted protocol, Megaface Challenge 1, MS-Celeb-1M Challenge 1\&2, explicitly encourage researchers to collect and clean a large-scale data set for enhancing the capability of deep neural network. Although data engineering is a valuable problem to computer vision researchers, this protocol is more incline to the industry participants. As evidence, the leaderboards of these experiments are mostly occupied by the companies holding invincible hardwares and data scales. This phenomenon may not be beneficial for developments of new models in academic community.

\begin{figure}[htbp]
\centering
\includegraphics[width=9cm]{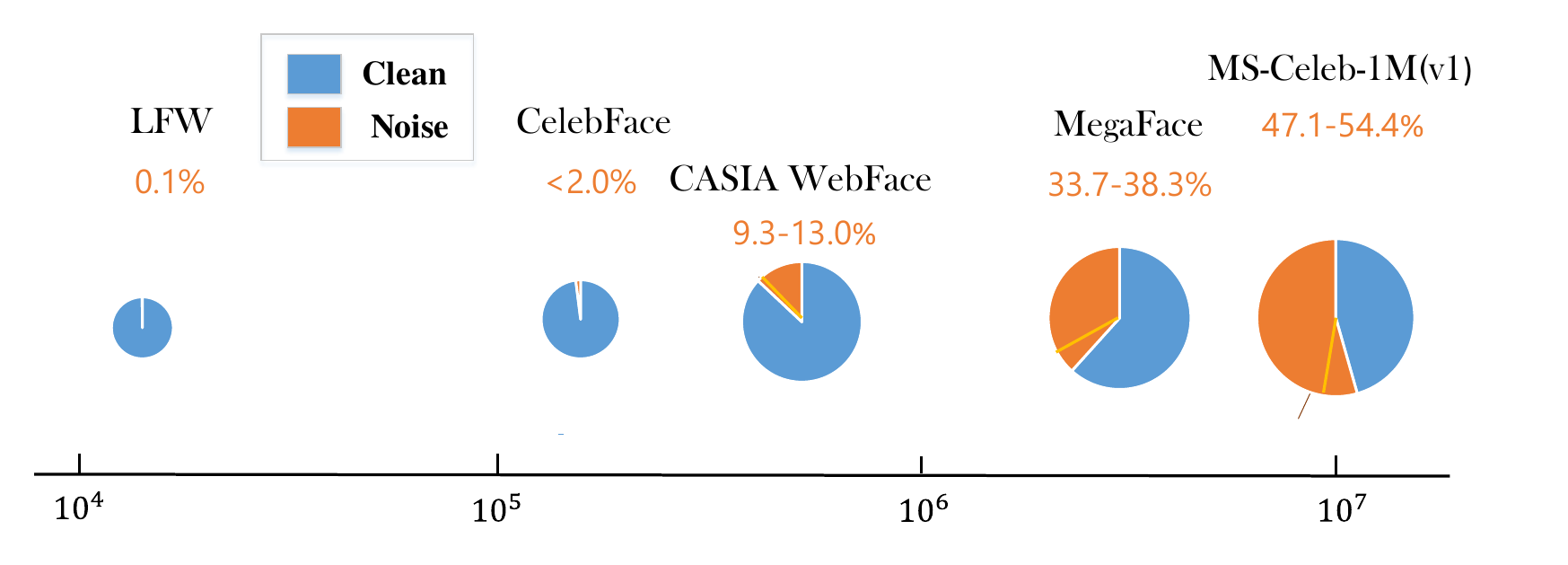}
\caption{ A visualization of the size and estimated noise percentage of datasets. \cite{Wang_2018_ECCV}}
\label{fig27}
\end{figure}

\textbf{Data noise.} Owing to data source and collecting strategies, existing large-scale datasets invariably contain label noises. Wang et al. \cite{Wang_2018_ECCV} profiled the noise distribution in existing datasets in Fig. \ref{fig27} and showed that the noise percentage increases dramatically along the scale of data. Moreover, they found that noise is more lethal on a 10,000-class problem of FR than on a 10-class problem of object classification and that label flip noise severely deteriorates the performance of a model, especially the model using A-softmax \cite{liu2017sphereface}. Therefore, building a sufficiently large and clean dataset for academic research is very meaningful. Deng et al. \cite{deng2019arcface} found there are serious label noise in MS-Celeb-1M \cite{guo2016ms}, and they cleaned the noise of MS-Celeb-1M, and made the refined dataset public available. Microsoft and Deepglint jointly released the largest public data set \cite{MS3} with cleaned labels, which includes 4M images cleaned from MS-Celeb-1M dataset and 2.8M aligned images of 100K Asian celebrities. Moreover, Zhan et al. \cite{Zhan_2018_ECCV} shifted the focus from cleaning the datasets to leveraging more unlabeled data. Through automatically assigning pseudo labels to unlabeled data with the help of relational graphs, they obtained competitive or even better results over the fully-supervised counterpart.

\textbf{Data bias.} 
Large-scale training datasets, such as CASIA-WebFace \cite{yi2014learning}, VGGFace2 \cite{cao2018vggface2} and MS-Celeb-1M \cite{guo2016ms}, are typically constructed by scraping websites like Google Images, and consist of celebrities on formal occasions: smiling, make-up, young, and beautiful. They are largely different from databases captured in the daily life (e.g. Megaface). The biases can be attributed to many exogenous factors in data collection, such as cameras, lightings, preferences over certain types of backgrounds, or annotator tendencies. Dataset biases adversely affect cross-dataset generalization; that is, the performance of the model trained on one dataset drops significantly when applied to another one. One persuasive evidence is presented by P.J. Phillips' study \cite{phillips2017cross} which conducted a cross benchmark assessment of VGGFace model \cite{parkhi2015deep} for face recognition. The VGGFace model achieves 98.95\% on LFW \cite{huang2007labeled} and 97.30\% on YTF \cite{wolf2011face}, but only obtains 26\%, 52\% and 85\% on Ugly, Bad and Good partition of GBU database \cite{phillips2012good}. 

Demographic bias (e.g., race/ethnicity, gender, age) in datasets is a universal but urgent issue to be solved in data bias field. In existing training and testing datasets, the male, White, and middle-aged cohorts always appear more frequently, as shown in Table \ref{race_ratio}, which inevitably causes deep learning models to replicate and even amplify these biases resulting in significantly different accuracies when deep models are applied to different demographic groups. 
Some researches \cite{wang2018deep,hupont2019demogpairs,serna2020sensitiveloss} showed that the female, Black, and younger cohorts are usually more difficult to recognize in FR systems trained with commonly-used datasets. For example, Wang et al. \cite{wang2019racial} proposed a Racial Faces in-the-Wild (RFW) database and proved that existing commercial APIs and the SOTA algorithms indeed work unequally for different races and the maximum difference in error rate between the best and worst groups is 12\%, as shown in Table \ref{race_bias}. Hupont et al. \cite{hupont2019demogpairs} showed that SphereFace has a TAR of 0.87 for White males which drops to 0.28 for Asian females, at a FAR of $1e-4$. Such bias can result in mistreatment of certain demographic groups, by either exposing them to a higher risk of fraud, or by making access to services more difficult. Therefore, addressing data bias and enhancing fairness of FR systems in real life are urgent and necessary tasks. Collecting balanced data to train a fair model or designing some debiasing algorithms are effective way. 

\begin{table*}[htbp]
	\begin{center}
    \footnotesize
    \setlength{\tabcolsep}{1.5mm}{
	\begin{tabular}{c|c|cccc|cc}
		\hline
       Train/ & \multirow{2}{*}{Database} & \multicolumn{4}{c|}{Race (\%)} & \multicolumn{2}{c}{Gender (\%)}  \\
       Test& & Caucasian & Asian & Indian & African& Female & Male \\ \hline \hline
       \multirow{3}{*}{train} & CASIA-WebFace \cite{yi2014learning} & 84.5&2.6 &1.6 &11.3 &41.1 & 58.9 \\
                                 & VGGFace2 \cite{cao2018vggface2} &74.2 &6.0 & 4.0&15.8  &40.7 &59.3\\
                                 & MS-Celeb-1M \cite{guo2016ms} &76.3&6.6 &2.6 &14.5 & - & -\\ \hline
        \multirow{2}{*}{test} & LFW \cite{huang2007labeled} &69.9 & 13.2& 2.9& 14.0 &25.8 &74.2  \\
                                 & IJB-A \cite{klare2015pushing} & 66.0& 9.8& 7.2& 17.0 & - & - \\ \hline
	\end{tabular}}
    \end{center}
    \caption{Statistical demographic information of commonly-used training and testing databases. \cite{wang2019racial,hupont2019demogpairs}}
    \label{race_ratio}
\end{table*}

\begin{table}[htbp]
	\begin{center}
    \footnotesize
    \setlength{\tabcolsep}{1.5mm}{
	\begin{tabular}{c|c|cccc}
		\hline
        \multirow{2}{*}{Model} &\multirow{2}{*}{LFW} & \multicolumn{4}{c}{RFW} \\
                               &                     & Caucasian & Indian & Asian & African \\ \hline \hline
         Microsoft  &98.22 &87.60 &82.83 &79.67 &75.83   \\
         Face++  & 97.03& 93.90& 88.55& 92.47& 87.50\\
         Baidu  &98.67 &89.13 & 86.53& 90.27&77.97  \\
         Amazon  &98.50 &90.45 &87.20 &84.87 &86.27  \\ \hline
         mean & 98.11 & 90.27 & 86.28 & 86.82 & 81.89 \\ \hline  \hline
         Center-loss \cite{wen2016discriminative} &98.75 & 87.18 &81.92 & 79.32&78.00 \\
         Sphereface \cite{liu2017sphereface} &99.27 & 90.80 &87.02 &82.95 &82.28  \\
         Arcface\tnote{1} \cite{deng2019arcface} & 99.40 & 92.15 &88.00 &83.98 &84.93  \\
         VGGface2\tnote{2} \cite{cao2018vggface2} & 99.30 &89.90 &86.13 &84.93 &83.38  \\ \hline
         mean & 99.18 & 90.01 & 85.77 & 82.80 & 82.15 \\ \hline
	\end{tabular}}
    \end{center}
    \caption{Racial bias in existing commercial recognition APIs and face recognition algorithms. Face verification accuracies (\%) on RFW database are given \cite{wang2019racial}. }
    \label{race_bias}
\end{table}

\subsection{Training protocols}

\begin{table*}[htbp]
\footnotesize
\centering
\caption{Performance of state of the arts on Megaface dataset}
\setlength{\tabcolsep}{0.5mm}{
\begin{tabular}{c|c|c|c|c|c|c|c|c|c}
	\hline
      \multirow{4}{*}{Method  }& \multicolumn{4}{c|}{Megaface challenge1}&\multirow{4}{*}{Method  } & \multicolumn{4}{|c}{Megaface challenge2}\\ \cline{2-5}\cline{7-10}
      & \multicolumn{2}{c|}{FaceScrub}&\multicolumn{2}{|c|}{FGNet}& &\multicolumn{2}{|c}{FaceScrub}&\multicolumn{2}{|c}{FGNet}\\ \cline{2-5}\cline{7-10}
	 &\tabincell{c}{Rank1\\@$10^{6}$} & \tabincell{c}{TPR\\@$10^{-6}$FPR }&\tabincell{c}{Rank1\\@$10^{6}$} & \tabincell{c}{TPR\\@$10^{-6}$FPR}& &\tabincell{c}{Rank1\\@$10^{6}$ }& \tabincell{c}{TPR\\@$10^{-6}$FPR}&\tabincell{c}{Rank1\\@$10^{6}$ }& \tabincell{c}{TPR\\@$10^{-6}$FPR }\\ \hline  \hline
     Arcface \cite{deng2019arcface}  & 0.9836 &0.9848&-&-&Cosface \cite{wang2018additive}& 0.7707&0.9030& 0.6118&0.6350\\ \hline
     Cosface \cite{wang2018additive} & 0.9833 &0.9841&-&-&&&&&\\ \hline
     A-softmax \cite{liu2017sphereface} &0.9743 &0.9766&-&-&&&&&\\ \hline
     Marginal loss \cite{deng2017marginal}&0.8028 &0.9264&0.6643& 0.4370&&&&&\\ \hline
\end{tabular}}
\label{megaface}
\end{table*}

\begin{table*}[htbp]
\footnotesize
\centering
\caption{Performance of state of the arts on MS-celeb-1M dataset}
\setlength{\tabcolsep}{0.5mm}{
\begin{tabular}{c|c|c|c|c|c|c|c}
	\hline
       \multirow{3}{*}{Method  }& \multicolumn{3}{|c|}{MS-celeb-1M challenge1}&  \multirow{3}{*}{Method  } & \multicolumn{3}{|c}{MS-celeb-1M challenge2}\\  \cline{2-4}\cline{6-8}
	& \tabincell{c}{External\\ Data} &\tabincell{c}{C@P=0.95\\random set} & \tabincell{c}{C@P=0.95\\hard set} &&  \tabincell{c}{External\\ Data} & \tabincell{c}{Top 1 Accuracy\\base set} &\tabincell{c}{C@P=0.99\\novel set} \\ \hline  \hline
     MCSM \cite{Xu2017High}& w & 0.8750 &0.7910 & Cheng et al. \cite{cheng2017know} &w&0.9974&0.9901\\ \hline
     Wang et al. \cite{8265435}&w/o&0.7500&0.6060& Ding et al. \cite{Ding2018Generative} &w/o&-&0.9484\\ \hline
     &&&&Hybrid Classifiers \cite{wu2017low}&w/o&0.9959&0.9264\\ \hline
     &&&&UP loss \cite{guo2017one}&w/o&0.9980&0.7748\\ \hline
\end{tabular}}
\label{tab7}
\end{table*}

\begin{table*}[htbp]
\footnotesize
\centering
\caption{Face Identification and Verification Evaluation of state of the arts on IJB-A dataset}
\setlength{\tabcolsep}{0.7mm}{
\begin{tabular}{c|ccc|cccc}
	\hline
      \multirow{2}{*}{Method  }& \multicolumn{3}{|c|}{IJB-A Verification (TAR@FAR)} & \multicolumn{4}{|c}{IJB-A Identification}\\  \cline{2-8}
	& 0.001 & 0.01 & 0.1 & FPIR=0.01 & FPIR=0.1 & Rank=1 & Rank=10 \\ \hline  \hline
     TDFF \cite{xiong2017good} &0.979$\pm$0.004 & 0.991$\pm$0.002 & 0.996$\pm$0.001 & 0.946$\pm$0.047 & 0.987$\pm$0.003 & 0.992$\pm$0.001 & 0.998$\pm$0.001 \\ \hline
     L2-softmax \cite{ranjan2017l2} & 0.943$\pm$0.005 & 0.970$\pm$0.004 & 0.984$\pm$0.002 & 0.915$\pm$0.041 & 0.956$\pm$0.006 & 0.973$\pm$0.005 & 0.988$\pm$0.003\\ \hline
     DA-GAN \cite{zhao2017dual}&0.930$\pm$0.005&0.976$\pm$0.007&0.991$\pm$0.003&0.890$\pm$0.039&0.949$\pm$0.009&0.971$\pm$0.007&0.989$\pm$0.003\\ \hline
     VGGface2 \cite{cao2018vggface2}&0.921$\pm$0.014 & 0.968$\pm$0.006 & 0.990$\pm$0.002&  0.883$\pm$0.038 & 0.946$\pm$0.004 & 0.982$\pm$0.004 &  0.994$\pm$0.001\\ \hline
     TDFF \cite{xiong2017good}&0.919$\pm$0.006 &0.961$\pm$0.007 &0.988$\pm$0.003 &0.878$\pm$0.035 &0.941$\pm$0.010 &0.964$\pm$0.006& 0.992$\pm$0.002\\ \hline
     NAN \cite{yang2017neural} & 0.881$\pm$0.011& 0.941$\pm$0.008& 0.979$\pm$0.004 &0.817$\pm$0.041& 0.917$\pm$0.009 &0.958$\pm$0.005& 0.986$\pm$0.003\\ \hline
     All-In-One Face \cite{ranjan2017all} & 0.823$\pm$0.020 &0.922$\pm$0.010 &0.976$\pm$0.004 &0.792$\pm$0.020 &0.887$\pm$0.014 &0.947$\pm$0.008 &0.988$\pm$0.003\\ \hline
     Template Adaptation \cite{crosswhite2017template} &0.836$\pm$0.027& 0.939$\pm$0.013& 0.979$\pm$0.004 &0.774$\pm$0.049 &0.882$\pm$0.016& 0.928$\pm$0.010& 0.986$\pm$0.003\\ \hline
     TPE \cite{sankaranarayanan2016}& 0.813$\pm$0.020 &0.900$\pm$0.010& 0.964$\pm$0.005 &0.753$\pm$0.030& 0.863$\pm$0.014 &0.932$\pm$0.010 &0.977$\pm$0.005\\ \hline
\end{tabular}}
\label{tab8}
\end{table*}

\begin{figure*}[htbp]
\centering
\includegraphics[width=15cm]{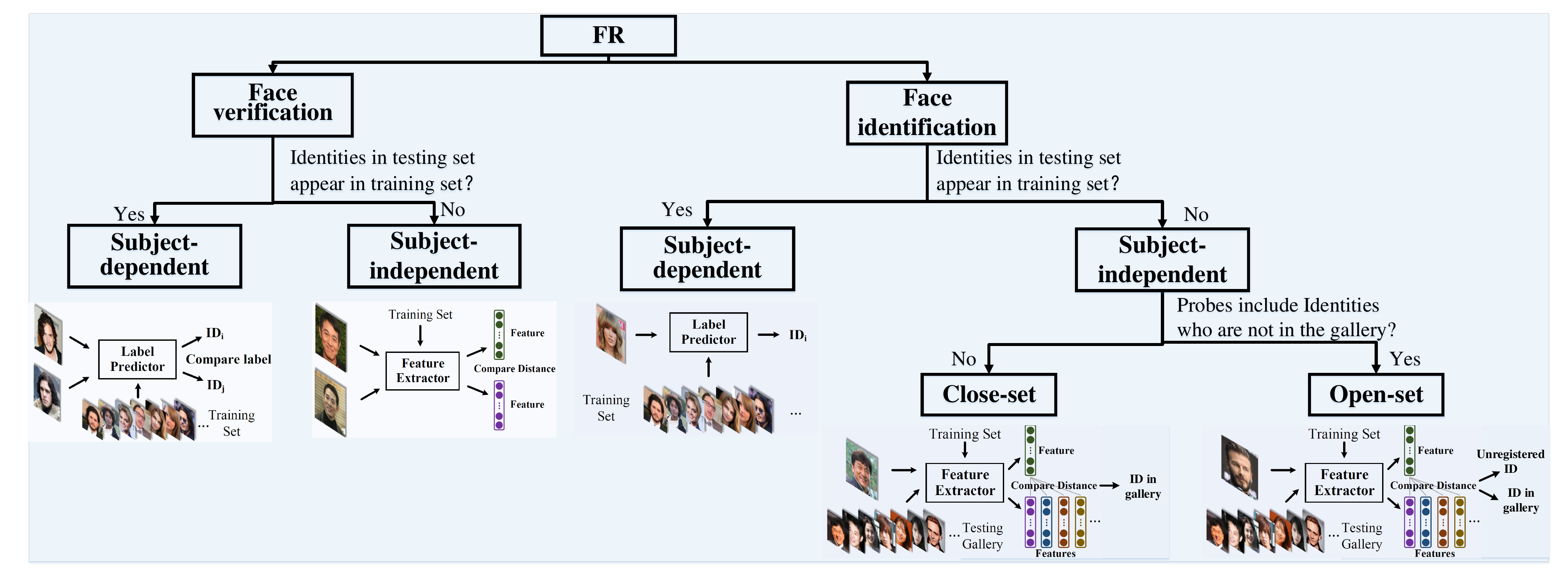}
\caption{ The comparison of different training protocol and evaluation tasks in FR. In terms of training protocol, FR can be classified into subject-dependent or subject-independent settings according to whether testing identities appear in training set. In terms of testing tasks, FR can be classified into face verification, close-set face identification, open-set face identification. }
\label{fig23}
\end{figure*}

In terms of training protocol, FR can be categorized into subject-dependent and subject-independent settings, as illustrated in Fig. \ref{fig23}.

\textbf{Subject-dependent protocol}. For subject-dependent protocol, all testing identities are predefined in training set, it is natural to classify testing face images to the given identities. Therefore, subject-dependent FR can be well addressed as a classification problem, where features are expected to be separable. The protocol is mostly adopted by the early-stage (before 2010) FR studies on FERET \cite{Phillips1998The}, AR \cite{martinez1998ar}, and is suitable only for some small-scale applications. The Challenge 2 of MS-Celeb-1M is the only large-scale database using subject-dependent training protocol.

\textbf{Subject-independent protocol}. For subject-independent protocol, the testing identities are usually disjoint from the training set, which makes FR more challenging yet close to practice. Because it is impossible to classify faces to known identities in training set, generalized representation is essential. Due to the fact that human faces exhibit similar intra-subject variations, deep models can display transcendental generalization ability when training with a sufficiently large set of generic subjects, where the key is to learn discriminative large-margin deep features. This generalization ability makes subject-independent FR possible. Almost all major face-recognition benchmarks, such as LFW \cite{huang2007labeled}, PaSC \cite{beveridge2013challenge}, IJB-A/B/C \cite{klare2015pushing,Whitelam2017IARPA,maze2018iarpa} and Megaface \cite{kemelmacher2016megaface,nech2017level}, require the tested models to be trained under subject-independent protocol.

\subsection{Evaluation tasks and performance metrics}

In order to evaluate whether our deep models can solve the different problems of FR in real life, many testing datasets are designed to evaluate the models in different tasks, i.e. face verification, close-set face identification and open-set face identification. In either task, a set of known subjects is initially enrolled in the system (the gallery), and during testing, a new subject (the probe) is presented. Face verification computes one-to-one similarity between the gallery and probe to determine whether the two images are of the same subject, whereas face identification computes one-to-many similarity to determine the specific identity of a probe face. When the probe appears in the gallery identities, this is referred to as closed-set identification; when the probes include those who are not in the gallery, this is open-set identification.

\begin{sidewaystable*}[htbp]
\scriptsize
\centering
\begin{threeparttable}
\caption{The commonly used FR datasets for testing}
\setlength{\tabcolsep}{0.7mm}{
 \begin{tabular}{c|c|c|c|c|c|c|c}
  \hline
   Datasets                       & \tabincell{l}{Publish\\ Time}   & \#photos        & \#subjects  &\tabincell{l}{\# of photos \\per subject \tnote{1}} & Metrics & Typical Methods \& Accuracy \tnote{2} & Key Features (Section)\\ \hline  \hline
   LFW \cite{huang2007labeled}                               & 2007  & 13K             & 5K          & 1/2.3/530 & \tabincell{l}{1:1: Acc, TAR vs. FAR (ROC);\\1:N: Rank-N, DIR vs. FAR (CMC)}  & \tabincell{l}{99.78\% Acc \cite{ranjan2017l2}; 99.63\% Acc \cite{schroff2015facenet}}         &  annotation with several attribute \\ \hline
   \tabincell{l}{MS-Celeb-1M \\ Challenge 1 \cite{guo2016ms}} & 2016  & 2K           & 1K	 & 2     &Coverage@P=0.95 & \tabincell{l}{random set: 87.50\%@P=0.95 \\hard set: 79.10\%@P=0.95 \cite{Xu2017High}}; & large-scale \\ \hline
   \tabincell{l}{MS-Celeb-1M \\ Challenge 2 \cite{guo2016ms}} & 2016  & \tabincell{l}{100K(base set) \\20K(novel set)} & \tabincell{l}{20K(base set)\\1K(novel set)}	 & 5/-/20     & Coverage@P=0.99 & 99.01\%@P=0.99 \cite{cheng2017know}& low-shot learning (\ref{Low-Shot}) \\ \hline
   \tabincell{l}{MS-Celeb-1M \\ Challenge 3 \cite{MS3}} & 2018  & \tabincell{l}{274K(ELFW) \\1M(DELFW)} & \tabincell{l}{5.7K(ELFW)\\1.58M(DELFW)}	 & -     & \tabincell{l}{1:1: TPR@FPR=1e-9;\\1:N: TPR@FPR=1e-3} & \tabincell{l}{1:1: 46.15\% \cite{deng2019arcface}\\1:N: 43.88\% \cite{deng2019arcface}}& trillion pairs; large distractors \\ \hline
   MegaFace	\cite{kemelmacher2016megaface,nech2017level}     & 2016  & 1M	& 690,572	 & 1.4   &\tabincell{l}{1:1: TPR vs. FPR (ROC);\\1:N: Rank-N (CMC)}& \tabincell{l}{1:1: 86.47\%@$10^{-6}$FPR \cite{schroff2015facenet};\\1:N: 70.50\% Rank-1 \cite{schroff2015facenet}} & large-scale; 1 million distractors\\ \hline
   IJB-A \cite{klare2015pushing}	                         & 2015  & 25,809	       & 500	     & 51.6     &\tabincell{l}{1:1: TAR vs. FAR (ROC);\\1:N: Rank-N, TPIR vs. FPIR (CMC, DET)}  &\tabincell{l}{1:1: 92.10\%@$10^{-3}$FAR \cite{cao2018vggface2};\\1:N: 98.20\% Rank-1 \cite{cao2018vggface2}} & cross-pose; template-based (\ref{Cross-Pose} and \ref{Set/Template})\\ \hline
   IJB-B \cite{Whitelam2017IARPA} &2017&  \tabincell{l}{11,754 images\\ 7,011 videos}& 1,845 & 41.6 & \tabincell{l}{1:1: TAR vs. FAR (ROC);\\1:N: Rank-N, TPIR vs. FPIR (CMC, DET)}  &\tabincell{l}{1:1: 52.12\%@$10^{-6}$FAR \cite{Kim_2020_CVPR};\\1:N: 90.20\% Rank-1 \cite{cao2018vggface2}} & cross-pose; template-based (\ref{Cross-Pose} and \ref{Set/Template})\\ \hline
    IJB-C \cite{maze2018iarpa} &2018&  \tabincell{l}{31.3K images\\11,779 videos}& 3,531 & 42.1 & \tabincell{l}{1:1: TAR vs. FAR (ROC);\\1:N: Rank-N, TPIR vs. FPIR (CMC, DET)}  &\tabincell{l}{1:1: 90.53\%@$10^{-6}$FAR \cite{Kim_2020_CVPR};\\1:N: 74.5\% Rank-1 \cite{tran2017disentangled}} & cross-pose; template-based (\ref{Cross-Pose} and \ref{Set/Template})\\ \hline
   RFW \cite{wang2019racial}                                 & 2018  & 40607            & 11429         &  3.6       & 1:1: Acc, TAR vs. FAR (ROC)     & \tabincell{l}{ Caucasian: 92.15\% Acc; Indian: 88.00\% Acc; \\Asian: 83.98\% Acc; African: 84.93\% Acc \cite{liu2017sphereface}}  & evaluating race bias \\ \hline
   DemogPairs  \cite{hupont2019demogpairs}        & 2019 & 10.8K            & 800 & 18                    & 1:1: TAR vs. FAR (ROC)            &  \tabincell{l}{White male: 88\%; White female: 87\%@$10^{-4}$FAR; \\Black male: 55\%; Black female: 65\%@$10^{-4}$FAR \cite{liu2017sphereface}}              & evaluating race and gender bias \\ \hline

   CPLFW \cite{CPLFW}	                                     & 2017  & 11652            & 3968          &  2/2.9/3   &1:1: Acc, TAR vs. FAR (ROC)& \tabincell{l}{77.90\% Acc \cite{parkhi2015deep}}   & cross-pose (\ref{Cross-Pose})\\ \hline
   CFP \cite{sengupta2016frontal}	                         & 2016  & 7,000		   & 500	     & 14      &1:1: Acc, EER, AUC, TAR vs. FAR (ROC)& \tabincell{l}{Frontal-Frontal: 98.67\% Acc \cite{peng2017reconstruction};\\ Frontal-Profile: 94:39\% Acc \cite{yin2017multi}}   & frontal-profile (\ref{Cross-Pose})\\ \hline
   SLLFW \cite{deng2017fine}	                             & 2017  & 13K             & 5K         &2.3 &  1:1: Acc, TAR vs. FAR (ROC)     & \tabincell{l}{ 85.78\% Acc \cite{parkhi2015deep}; 78.78\% Acc \cite{taigman2014deepface}}   & fine-grained \\ \hline
   UMDFaces \cite{bansal2017umdfaces}                        & 2016  & 367,920         & 8,501       & 43.3     &  1:1: Acc, TPR vs. FPR (ROC)& 69.30\%@$10^{-2}$FAR \cite{krizhevsky2012imagenet}   & annotation with bounding boxes, 21 keypoints, gender and 3D pose  \\ \hline
   YTF \cite{wolf2011face}	                                 & 2011  & 3,425 	       & 1,595	     & 48/181.3/6,070  &1:1: Acc&  \tabincell{l}{97.30\% Acc \cite{parkhi2015deep}; 96.52\% Acc \cite{rao2017attention}}  & video (\ref{Video}) \\ \hline
   PaSC	\cite{beveridge2013challenge}                        & 2013  & 2,802 	       & 265	     & --	    &1:1: VR vs. FAR (ROC)& 95.67\%@$10^{-2}$FAR \cite{rao2017attention}& video (\ref{Video})\\ \hline
   YTC  \cite{kim2008face}                                   & 2008  & 1,910           & 47          & --   & 1:N: Rank-N (CMC) &  97.82\% Rank-1 \cite{rao2017attention}; 97.32\% Rank-1 \cite{rao2017learning}  & video (\ref{Video})\\ \hline
   CALFW \cite{zheng2017cross}	                             & 2017  & 12174             & 4025          & 2/3/4     &1:1: Acc, TAR vs. FAR (ROC)& \tabincell{l}{ 86.50\% Acc \cite{parkhi2015deep}; 82.52\% Acc \cite{chen2017noisy}}    & cross-age; 12 to 81 years old (\ref{Cross-Age})\\ \hline
   MORPH  \cite{ricanek2006morph}                            & 2006  & 55,134          & 13,618      & 4.1    & 1:N: Rank-N (CMC) & 94.4\% Rank-1 \cite{Lin2016Cross} & cross-age, 16 to 77 years old (\ref{Cross-Age})\\ \hline
   CACD \cite{chen2014cross}                                 & 2014  & 163,446         & 2000        & 81.7    & \tabincell{l}{1:1 (CACD-VS): Acc, TAR vs. FAR (ROC)\\1:N: MAP} & \tabincell{l}{1:1 (CACD-VS): 98.50\% Acc \cite{wen2016latent}\\1:N: 69.96\% MAP (2004-2006)\cite{zheng2017age}}   & cross-age, 14 to 62 years old (\ref{Cross-Age})\\ \hline
   FG-NET \cite{FGNET}                                       & 2010  & 1,002            & 82          & 12.2    & 1:N: Rank-N (CMC)&  88.1\% Rank-1 \cite{wen2016latent}   & cross-age, 0 to 69 years old (\ref{Cross-Age}) \\ \hline
   \tabincell{l}{CASIA \\NIR-VIS v2.0 \cite{li2013casia}}    & 2013  & 17,580          & 725         & 24.2     & 1:1: Acc, VR vs. FAR (ROC) & 98.62\% Acc, 98.32\%@$10^{-3}$FAR \cite{wu2018coupled} & NIR-VIS; with eyeglasses, pose and expression variation (\ref{NIR-VIS})\\ \hline
   CASIA-HFB \cite{Li2009The}                               & 2009  & 5097             & 202        & 25.5     &1:1: Acc, VR vs. FAR (ROC)&97.58\% Acc, 85.00\%@$10^{-3}$FAR \cite{reale2016seeing}&NIR-VIS; with eyeglasses and expression variation (\ref{NIR-VIS})\\ \hline
   CUFS \cite{wang2009face}                                  & 2009  & 1,212           & 606         & 2     & 1:N: Rank-N (CMC)&  100\% Rank-1 \cite{zhang2015end}  & sketch-photo (\ref{Photo-Sketch}) \\ \hline
   CUFSF \cite{zhang2011coupled}                             & 2011  & 2,388           & 1,194       & 2     & 1:N: Rank-N (CMC)&  51.00\% Rank-1 \cite{wang2018high}  & sketch-photo; lighting variation; shape exaggeration (\ref{Photo-Sketch}) \\ \hline
   Bosphorus \cite{savran2008bosphorus}                      & 2008  & 4,652           & 105         & 31/44.3/54     &\tabincell{l}{1:1: TAR vs. FAR (ROC);\\1:N: Rank-N (CMC)} & 1:N: 99.20\% Rank-1 \cite{kim2017deep}  &  3D; 34 expressions, 4 occlusions and different poses (\ref{3D})\\ \hline
   BU-3DFE \cite{yin20063d}                                  & 2006  & 2,500           & 100         & 25      &\tabincell{l}{1:1: TAR vs. FAR (ROC);\\1:N: Rank-N (CMC)}&   1:N: 95.00\% Rank-1 \cite{kim2017deep}      & 3D; different expressions (\ref{3D})\\ \hline
   FRGCv2 \cite{phillips2005overview}                        & 2005  & 4,007           & 466         & 1/8.6/22     & \tabincell{l}{1:1: TAR vs. FAR (ROC);\\1:N: Rank-N (CMC)}&  1:N: 94.80\% Rank-1 \cite{kim2017deep} & 3D; different expressions (\ref{3D})\\ \hline
   Guo et al. \cite{guo2014face}                             & 2014  & 1,002           & 501         & 2       &1:1: Acc, TAR vs. FAR (ROC) & 94.8\% Rank-1, 65.9\%@$10^{-3}$FAR \cite{li2018anti}& make-up; female (\ref{Makeup})  \\ \hline
   FAM \cite{hu2013makeup}                                   & 2013  & 1,038           & 519         & 2      & 1:1: Acc, TAR vs. FAR (ROC) & 88.1\% Rank-1, 52.6\%@$10^{-3}$FAR \cite{li2018anti}  & make-up; female and male (\ref{Makeup})  \\ \hline
   CASIA-FASD \cite{Zhang2012A}                              & 2012  & 600             & 50          & 12    &EER, HTER   & 2.67\% EER, 2.27\% HTER \cite{atoum2017face}& anti-spoofing (\ref{anti-spoofing})  \\ \hline
   Replay-Attack \cite{Chingovska_BIOSIG-2012}               & 2012  & 1,300            & 50          & --    & EER, HTER  &0.79\% EER, 0.72\% HTER \cite{atoum2017face}& anti-spoofing (\ref{anti-spoofing})      \\ \hline
   WebCaricature \cite{huo2017webcaricature}                 & 2017  & 12,016           & 252         & --    & \tabincell{l}{1:1: TAR vs. FAR (ROC);\\1:N: Rank-N (CMC)}& \tabincell{l}{1:1: 34.94\%@$10^{-1}$FAR \cite{huo2017webcaricature};\\1:N: 55.41\% Rank-1 \cite{huo2017webcaricature}} & Caricature (\ref{Photo-Sketch})  \\ \hline
 \end{tabular}}
 \label{tab5}
 \begin{tablenotes}
  \item[1] The min/average/max numbers of photos or frames per subject
  \item[2] We only present the typical methods that are published in a paper, and the accuracies of the most challenging scenarios are given.
\end{tablenotes}
 \end{threeparttable}
\end{sidewaystable*}

\textbf{Face verification} is relevant to access control systems, re-identification, and application independent evaluations of FR algorithms. It is classically measured using the receiver operating characteristic (ROC) and estimated mean accuracy (Acc). At a given threshold (the independent variable), ROC analysis measures the true accept rate (TAR), which is the fraction of genuine comparisons that correctly exceed the threshold, and the false accept rate (FAR), which is the fraction of impostor comparisons that incorrectly exceed the threshold. And Acc is a simplified metric introduced by LFW \cite{huang2007labeled}, which represents the percentage of correct classifications. With the development of deep FR, more accurate recognitions are required. Customers concern more about the TAR when FAR is kept in a very low rate in most security certification scenario. PaSC \cite{beveridge2013challenge}  reports TAR at a FAR of $10^{-2}$; IJB-A \cite{klare2015pushing} evaluates TAR at a FAR of $ 10^{-3}$; Megaface \cite{kemelmacher2016megaface,nech2017level} focuses on TAR@$ 10^{-6}$FAR; especially, in MS-celeb-1M challenge 3 \cite{MS3}, TAR@$ 10^{-9}$FAR is reported.

\textbf{Close-set face identification} is relevant to user driven searches (e.g., forensic identification), rank-N and cumulative match characteristic (CMC) is commonly used metrics in this scenario. Rank-N is based on what percentage of probe searches return the probe's gallery mate within the top $k$ rank-ordered results. The CMC curve reports the percentage of probes identified within a given rank (the independent variable). IJB-A/B/C \cite{klare2015pushing,Whitelam2017IARPA,maze2018iarpa} concern on the rank-1 and rank-5 recognition rate. The MegaFace challenge \cite{kemelmacher2016megaface,nech2017level} systematically evaluates rank-1 recognition rate function of increasing number of gallery distractors (going from 10 to 1 Million), the results of the SOTA evaluated on MegaFace challenge are listed in Table \ref{megaface}. Rather than rank-N and CMC, MS-Celeb-1M \cite{guo2016ms} further applies a precision-coverage curve to measure identification performance under a variable threshold $t$. The probe is rejected when its confidence score is lower than $t$. The algorithms are compared in term of what fraction of passed probes, i.e. coverage, with a high recognition precision, e.g. 95\% or 99\%, the results of the SOTA evaluated on MS-Celeb-1M challenge are listed in Table \ref{tab7}.

\textbf{Open-set face identification} is relevant to high throughput face search systems (e.g., de-duplication, watch list identification), where the recognition system should reject unknown/unseen subjects (probes who do not present in gallery) at test time. At present, there are very few databases covering the task of open-set FR. IJB-A/B/C \cite{klare2015pushing,Whitelam2017IARPA,maze2018iarpa} benchmarks introduce a decision error tradeoff (DET) curve to characterize the the false negative identification rate (FNIR) as function of the false positive identification rate (FPIR). FPIR measures what fraction of comparisons between probe templates and non-mate gallery templates result in a match score exceeding $T$. At the same time, FNIR measures what fraction of probe searches will fail to match a mated gallery template above a score of $T$. The algorithms are compared in term of the FNIR at a low FPIR, e.g. 1\% or 10\%, the results of the SOTA evaluated on IJB-A dataset as listed in Table \ref{tab8}.

\subsection{Evaluation Scenes and Data}

\begin{figure*}[htbp]
\centering
\includegraphics[width=\textwidth]{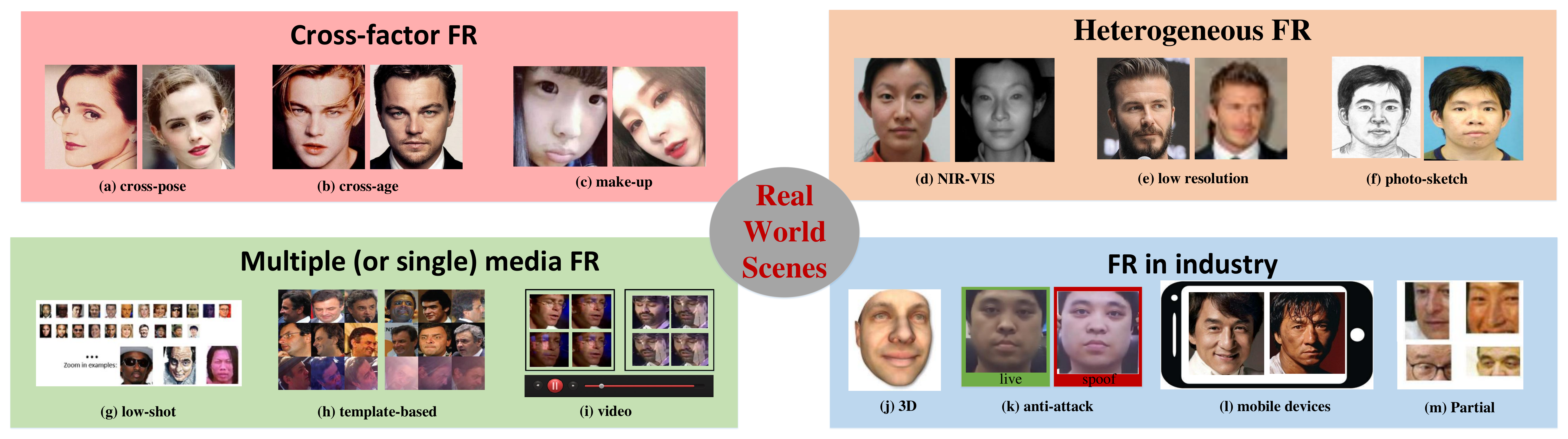}
\caption{ The different scenes of FR. We divide FR scenes into four categories: cross-factor FR, heterogenous FR, multiple (or single) media FR and FR in industry. There are many testing datasets and special FR methods designed for each scene.}
\label{fig15}
\end{figure*}

Public available training databases are mostly collected from the photos of celebrities due to privacy issue, it is far from images captured in the daily life with diverse scenes. In order to study different specific scenarios, more difficult and realistic datasets are constructed accordingly, as shown in Table \ref{tab5}. According to their characteristics, we divide these scenes into four categories: cross-factor FR, heterogenous FR, multiple (or single) media FR and FR in industry (Fig. \ref{fig15}).

\begin{itemize}
  \item Cross-factor FR. Due to the complex nonlinear facial appearance, some variations will be caused by people themselves, such as cross-pose, cross-age, make-up, and disguise. For example, CALFW \cite{zheng2017cross}, MORPH  \cite{ricanek2006morph}, CACD \cite{chen2014cross} and FG-NET \cite{FGNET} are commonly used datasets with different age range; CFP \cite{sengupta2016frontal} only focuses on frontal and profile face, CPLFW \cite{CPLFW} is extended from LFW and contains different poses. Disguised faces in the wild (DFW) \cite{kushwaha2018disguised} evaluates face recognition across disguise.
  \item Heterogenous FR. It refers to the problem of matching faces across different visual domains. The domain gap is mainly caused by sensory devices and cameras settings, e.g. visual light vs. near-infrared and photo vs. sketch. For example, CUFSF \cite{zhang2011coupled} and CUFS \cite{wang2009face} are commonly used photo-sketch datasets and CUFSF dataset is harder due to lighting variation and shape exaggeration.
  \item Multiple (or single) media FR. Ideally, in FR, many images of each subject are provided in training datasets and image-to-image recognitions are performed when testing. But the situation will be different in reality. Sometimes, the number of images per person in training set could be very small, such as MS-Celeb-1M challenge 2 \cite{guo2016ms}. This challenge is often called low- shot or few-shot FR. Moreover, each subject face in test set may be enrolled with a set of images and videos and set-to-set recognition should be performed, such as IJB-A \cite{klare2015pushing} and PaSC \cite{beveridge2013challenge}.
  \item FR in industry. Although deep FR has achieved beyond human performance on some standard benchmarks, but some other factors should be given more attention rather than accuracy when deep FR is adopted in industry, e.g. anti-attack (CASIA-FASD \cite{Zhang2012A}) and 3D FR (Bosphorus \cite{savran2008bosphorus}, BU-3DFE \cite{yin20063d} and FRGCv2 \cite{phillips2005overview}). Compared to publicly available 2D face databases, 3D scans are hard to acquire, and the number of scans and subjects in public 3D face databases is still limited, which hinders the development of 3D deep FR.
\end{itemize}

\section{Diverse recognition scenes of deep learning}

Despite the high accuracy in the LFW \cite{huang2007labeled} and Megaface \cite{kemelmacher2016megaface,nech2017level} benchmarks, the performance of FR models still hardly meets the requirements in real-world application. A conjecture in industry is made that results of generic deep models can be improved simply by collecting big datasets of the target scene. However, this holds only to a certain degree. More and more concerns on privacy may make the collection and human-annotation of face data become illegal in the future. Therefore, significant efforts have been paid to design excellent algorithms to address the specific problems with limited data in these realistic scenes. In this section, we present several special algorithms of FR.

\subsection{Cross-Factor Face Recognition}

\subsubsection{Cross-Pose Face Recognition}\label{Cross-Pose}

As \cite{sengupta2016frontal} shows that many existing algorithms suffer a decrease of over 10\% from frontal-frontal to frontal-profile verification, cross-pose FR is still an extremely challenging scene. In addition to the aforementioned methods, including ``one-to-many augmentation'', ``many-to-one normalization'' and assembled networks (Section \ref{Pre-processing} and \ref{Multiple Network}), there are some other algorithms designed for cross-pose FR. Considering the extra burden of above methods, Cao et al. \cite{cao2018pose} attempted to perform frontalization in the deep feature space rather than the image space. A deep residual equivariant mapping (DREAM) block dynamically added residuals to an input representation to transform a profile face to a frontal image. Chen et al. \cite{chen2018deep} proposed to combine feature extraction with multi-view subspace learning to simultaneously make features be more pose-robust and discriminative. Pose Invariant Model (PIM) \cite{Zhao2018towardspose} jointly performed face frontalization and learned pose invariant representations end-to-end to allow them to mutually boost each other, and further introduced unsupervised cross-domain adversarial training and a °learning to learn strategy to provide high-fidelity frontal reference face images.

\subsubsection{Cross-Age Face Recognition}\label{Cross-Age}

\begin{figure}[htbp]
\centering
\includegraphics[width=9cm]{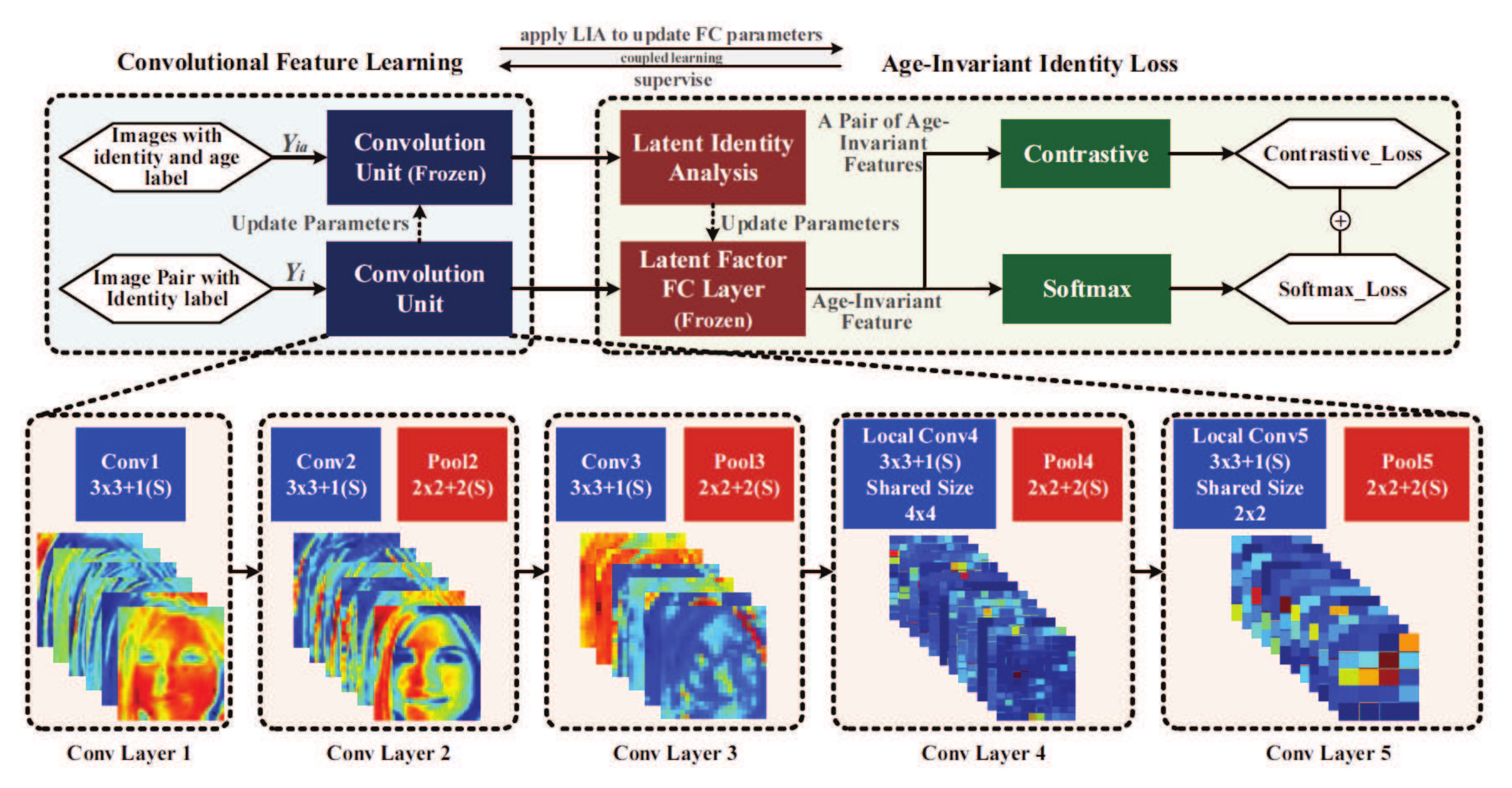}
\caption{ The architecture of the cross-age FR with LIA. \cite{wen2016latent}}
\label{fig12}
\end{figure}

Cross-age FR is extremely challenging due to the changes in facial appearance by the aging process over time. One direct approach is to synthesize the desired image with target age such that the recognition can be performed in the same age group. A generative probabilistic model was used by \cite{duong2017temporal} to model the facial aging process at each short-term stage. The identity-preserved conditional generative adversarial networks (IPCGANs) \cite{Wang2018faceaging} framework utilized a conditional-GAN to generate a face in which an identity-preserved module preserved the identity information and an age classifier forced the generated face with the target age. Antipov et al. \cite{antipov2017face} proposed to age faces by GAN, but the synthetic faces cannot be directly used for face verification due to its imperfect preservation of identities. Then, they used a local manifold adaptation (LMA) approach \cite{antipov2017boosting} to solve the problem of \cite{antipov2017face}. In \cite{yang2018learning}, high-level age-specific features conveyed by the synthesized face are estimated by a pyramidal adversarial discriminator at multiple scales to generate more lifelike facial details. An alternative to address the cross-age problem is to decompose aging and identity components separately and extract age-invariant representations. Wen et al. \cite{wen2016latent} developed a latent identity analysis (LIA) layer to separate these two components, as shown in Fig. \ref{fig12}. In \cite{zheng2017age}, age-invariant features were obtained by subtracting age-specific factors from the representations with the help of the age estimation task. In \cite{Wang_2018_ECCV}, face features are decomposed in the spherical coordinate system, in which the identity-related components are represented with angular coordinates and the age-related information is encoded with radial coordinate. Additionally, there are other methods designed for cross-age FR. For example, Bianco ett al. \cite{bianco2017large} and El et al. \cite{el2017age} fine-tuned the CNN to transfer knowledge across age. Wang et al. \cite{wang2017unleash} proposed a siamese deep network to perform multi-task learning of FR and age estimation. Li et al. \cite{li2018distance} integrated feature extraction and metric learning via a deep CNN.

\subsubsection{Makeup Face Recognition}\label{Makeup}

\begin{figure}[htbp]
\centering
\includegraphics[width=9cm]{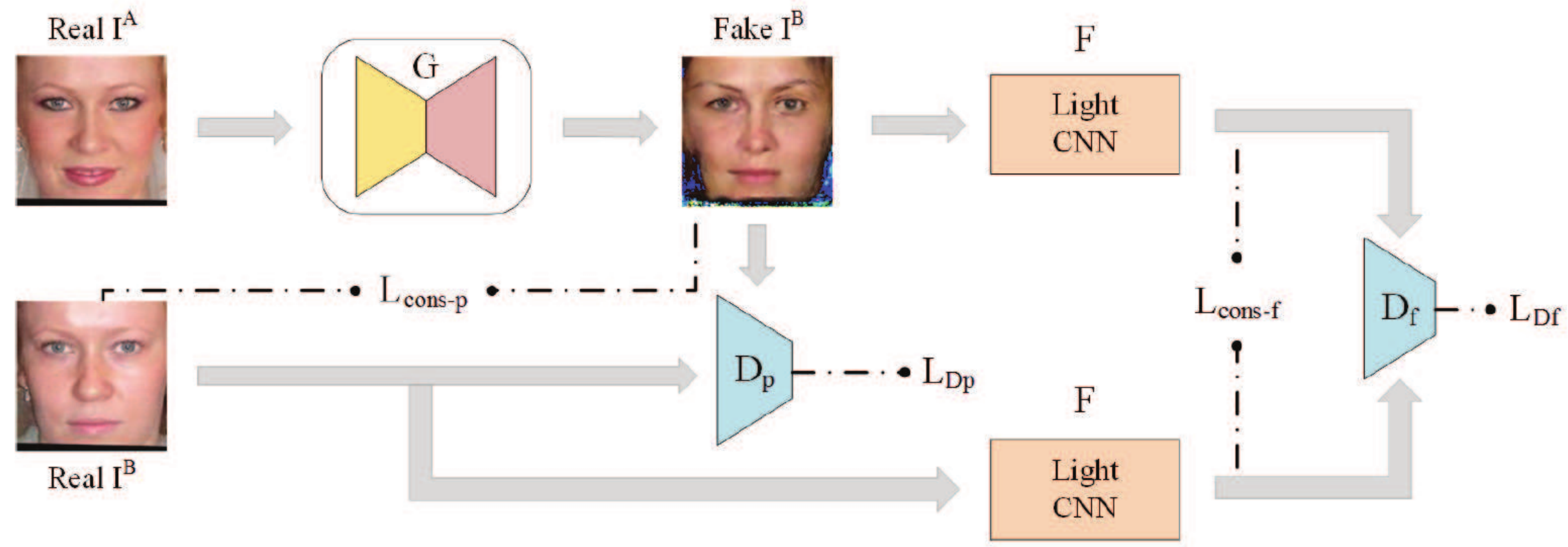}
\caption{ The architecture of BLAN. \cite{li2018anti}}
\label{fig14}
\end{figure}

Makeup is widely used by the public today, but it also brings challenges for FR due to significant facial appearance changes. The research on matching makeup and nonmakeup
face images is receiving increasing attention. Li et al. \cite{li2018anti} generated nonmakeup images from makeup ones by a bi-level adversarial network (BLAN) and then used the synthesized nonmakeup images for verification as shown in Fig. \ref{fig14}. Sun et al. \cite{sun2017weakly} pretrained a triplet network on videos and fine-tuned it on a small makeup datasets. Specially, facial disguise \cite{kushwaha2018disguised,singh2019disguised,singh2019recognizing} is a challenging research topic in makeup face recognition. By using disguise accessories such as wigs, beard, hats, mustache, and heavy makeup, disguise introduces two variations: (i) when a person wants to obfuscate his/her own identity, and (ii) another individual impersonates someone else's identity. Obfuscation increases intra-class variations whereas impersonation reduces the inter-class dissimilarity, thereby affecting face recognition/verification task. To address this issue, a variety of methods are proposed. Zhang et al. \cite{zhang2018deep} first trained two DCNNs for generic face recognition and then used Principal Components Analysis (PCA) to find the transformation matrix for disguised face recognition adaptation. Kohli et al. \cite{kohli2018face} finetuned models using disguised faces. Smirnov et al. \cite{smirnov2018hard} proposed a hard example mining method benefitted from class-wise (Doppelganger Mining \cite{smirnov2017doppelganger}) and example-wise mining to learn useful deep embeddings for disguised face recognition. Suri et al. \cite{suri2018matching} learned the representations of images in terms of colors, shapes, and textures (COST) using an unsupervised dictionary learning method, and utilized the combination of COST features and CNN features to perform recognition.

\subsection{Heterogenous Face Recognition}

\subsubsection{NIR-VIS Face Recognition}\label{NIR-VIS}

Due to the excellent performance of the near-infrared spectrum (NIS) images under low-light scenarios, NIS images are widely applied in surveillance systems. Because most enrolled databases consist of visible light (VIS) spectrum images, how to recognize a NIR face from a gallery of VIS images has been a hot topic. Saxena et al. \cite{saxena2016heterogeneous} and Liu et al. \cite{liu2016transferring} transferred the VIS deep networks to the NIR domain by fine-tuning. Lezama et al. \cite{lezama2017not} used a VIS CNN to recognize NIR faces by transforming NIR images to VIS faces through cross-spectral hallucination and restoring a low-rank structure for features through low-rank embedding. Reale et al. \cite{reale2016seeing} trained a VISNet (for visible images) and a NIRNet (for near-infrared images), and coupled their output features by creating a siamese network. He et al. \cite{he2017learning,he2018wasserstein} divided the high layer of the network into a NIR layer, a VIS layer and a NIR-VIS shared layer, then, a modality-invariant feature can be learned by the NIR-VIS shared layer. Song et al. \cite{song2018adversarial} embedded cross-spectral face hallucination and discriminative feature learning into an end-to-end adversarial network. In \cite{wu2018coupled}, the low-rank relevance and cross-modal ranking were used to alleviate the semantic gap.

\subsubsection{Low-Resolution Face Recognition}

Although deep networks are robust to low resolution to a great extent, there are still a few studies focused on promoting the performance of low-resolution FR. For example, Zangeneh et al. \cite{zangeneh2020low} proposed a CNN with a two-branch architecture (a super-resolution network and a feature extraction network) to map the high- and low-resolution face images into a common space where the intra-person distance is smaller than the inter-person distance. Shen et al. \cite{Shen2018dublurring} exploited the face semantic information and local structural constraints to better restore the shape and detail of face images. In addition, they optimized the network with perceptual and adversarial losses to produce photo-realistic results.

\subsubsection{Photo-Sketch Face Recognition}\label{Photo-Sketch}

\begin{figure}[htbp]
\centering
\includegraphics[width=9cm]{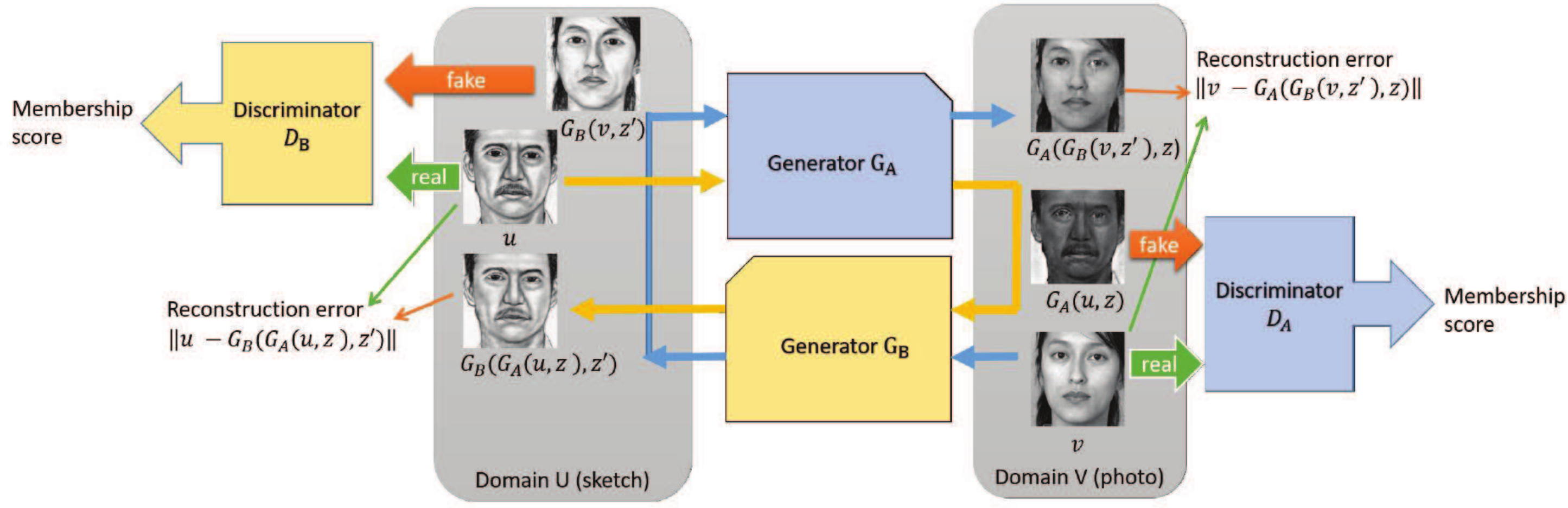}
\caption{ The architecture of DualGAN. \cite{yi2017dualgan}}
\label{fig16}
\end{figure}

The photo-sketch FR may help law enforcement to quickly identify suspects. The commonly used methods can be categorized as two classes. One is to utilize transfer learning to directly match photos to sketches. Deep networks are first trained using a large face database of photos and are then fine-tuned using small sketch database \cite{mittal2015composite,galea2017forensic}. The other is to use the image-to-image translation, where the photo can be transformed to a sketch or the sketch to a photo; then, FR can be performed in one domain. Zhang et al. \cite{zhang2015end} developed a fully convolutional network with generative loss and a discriminative regularizer to transform photos to sketches. Zhang et al. \cite{zhang2017content} utilized a branched fully convolutional neural network (BFCN) to generate a structure-preserved sketch and a texture-preserved sketch, and then they fused them together via a probabilistic method. Recently, GANs have achieved impressive results in image generation. Yi et al. \cite{yi2017dualgan}, Kim et al. \cite{kim2017learning} and Zhu et al. \cite{zhu2017unpaired} used two generators, $G_A$ and $G_B$, to generate sketches from photos and photos from sketches, respectively (Fig. \ref{fig16}). Based on \cite{zhu2017unpaired}, Wang et al. \cite{wang2018high} proposed a multi-adversarial network to avoid artifacts by leveraging the implicit presence of feature maps of different resolutions in the generator subnetwork. Similar to photo-sketch FR, photo-caricature FR is one kind of heterogenous FR scenes which is challenging and important to understanding of face perception. Huo et al. \cite{huo2017webcaricature} built a large dataset of caricatures and photos, and provided several evaluation protocols and their baseline performances for comparison.

\subsection{Multiple (or single) media Face Recognition}

\subsubsection{Low-Shot Face Recognition}\label{Low-Shot}

\begin{figure}[htbp]
\centering
\includegraphics[width=9cm]{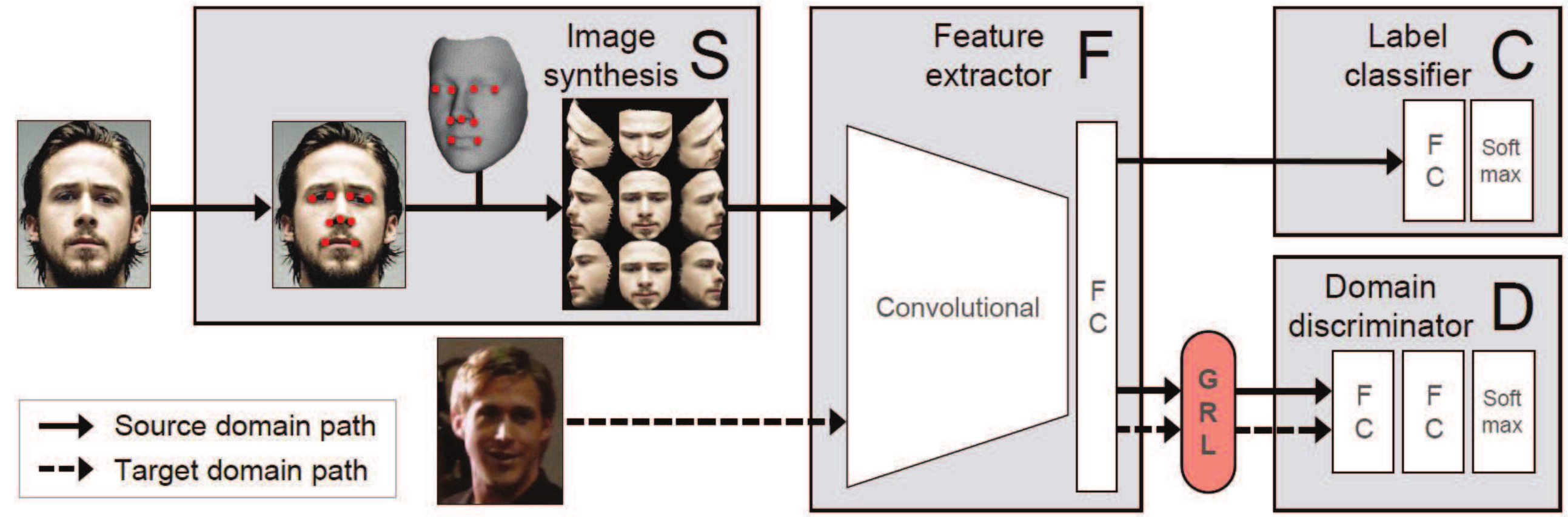}
\caption{ The architecture of a single sample per person domain adaptation network (SSPP-DAN). \cite{hong2017sspp}}
\label{fig13}
\end{figure}

For many practical applications, such as surveillance and security, the FR system should recognize persons with a very limited number of training samples or even with only one sample. The methods of low-shot learning can be categorized as 1) synthesizing training data and 2) learning more powerful features. Hong et al. \cite{hong2017sspp} generated images in various poses using a 3D face model and adopted deep domain adaptation to handle other variations, such as blur, occlusion, and expression (Fig. \ref{fig13}). Choe et al. \cite{choe2017face} used data augmentation methods and a GAN for pose transition and attribute boosting to increase the size of the training dataset. Wu et al. \cite{wu2017low} proposed a framework with hybrid classifiers using a CNN and a nearest neighbor (NN) model. Guo et al. \cite{guo2017one} made the norms of the weight vectors of the one-shot classes and the normal classes aligned to address the data imbalance problem. Cheng et al. \cite{cheng2017know} proposed an enforced softmax that contains optimal dropout, selective attenuation, L2 normalization and model-level optimization. Yin et al. \cite{yin2019feature} augmented feature space of low-shot classes by transferring the principal components from regular to low-shot classes to encourage the variance of low-shot classes to mimic that of regular classes.

\subsubsection{Set/Template-Based Face Recognition}\label{Set/Template}

Different from traditional image-to-image recognition, set-to-set recognition takes a set (heterogeneous contents containing both images and videos) as the smallest unit of representation. This kind of setting does reflect the real-world biometric scenarios, thereby attracting a lot of attention. After learning face representations of media in each set, two strategies are generally adopted to perform set-to-set matching. One is to use these representations to perform pair-wise similarity comparison of two sets and aggregate the results into a single and final score by max score pooling \cite{masi2016pose}, average score pooling \cite{lu2015multi} and its variations \cite{zhao2017unconstrained,bodla2017deep}. The other strategy is feature pooling \cite{masi2016pose,chen2016unconstrained,sankaranarayanan2016} which first aggregates face representations into a single representation for each set and then performs a comparison between two sets. In addition to the commonly used strategies, there are also some novel methods proposed for set/template-based FR. For example, Hayat et al. \cite{hayat2014learning} proposed a deep heterogeneous feature fusion network to exploit the features' complementary information generated by different CNNs. Liu et al. \cite{Liu_2018_ECCV} introduced the actor-critic reinforcement learning for set-based FR. They casted the inner-set dependency modeling to a Markov decision process in the latent space, and trained a dependency-aware attention control agent to make attention control for each image in each step.

\subsubsection{Video Face Recognition}\label{Video}

\begin{figure}[htbp]
\centering
\includegraphics[width=8cm]{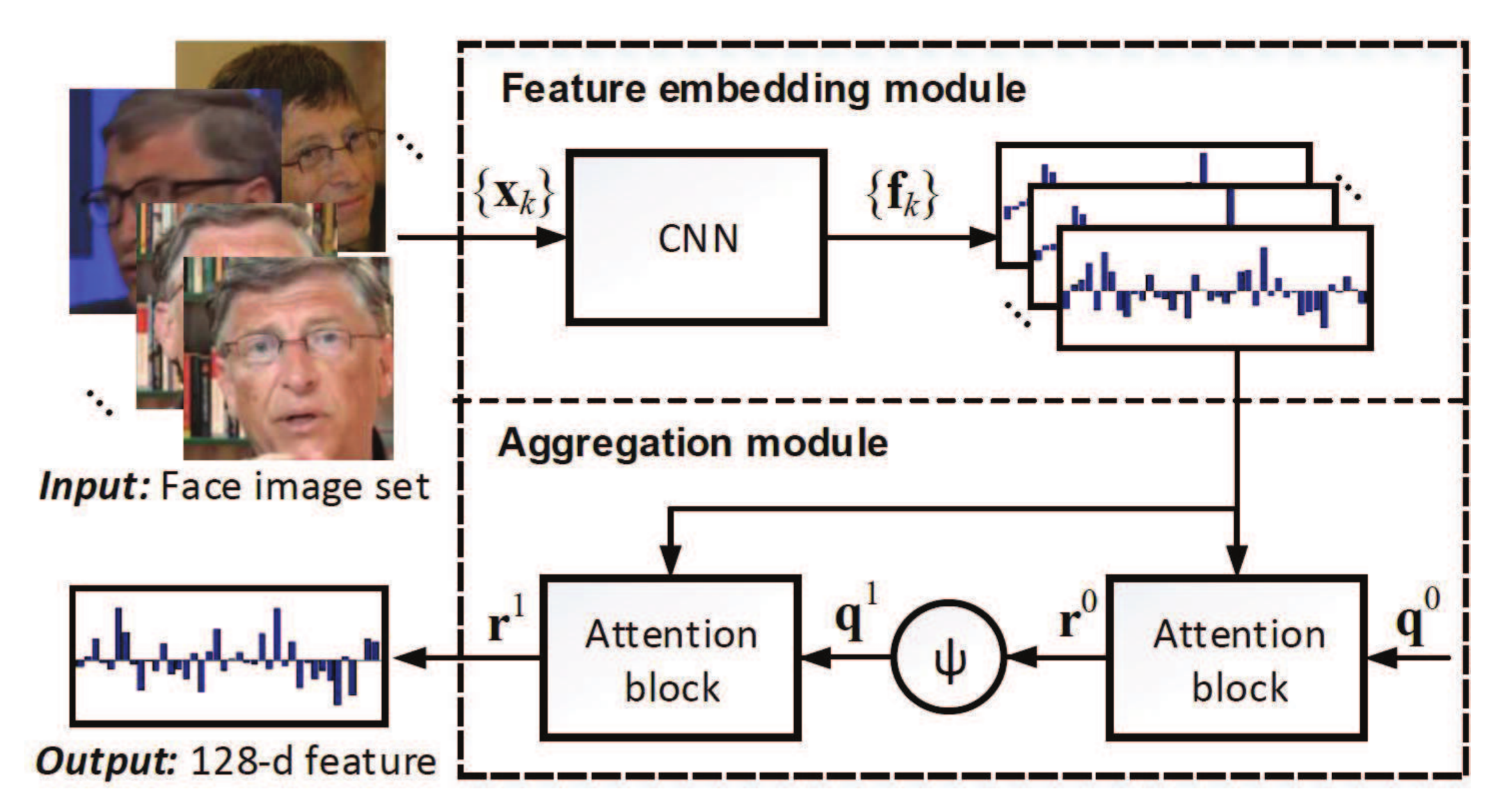}
\caption{ The FR framework of NAN. \cite{yang2017neural}}
\label{fig11}
\end{figure}

There are two key issues in video FR: one is to integrate the information across different frames together to build a representation of the video face, and the other is to handle video frames with severe blur, pose variations, and occlusions. For frame aggregation, Yang et al. \cite{yang2017neural} proposed a neural aggregation network (NAN) in which the aggregation module, consisting of two attention blocks driven by a memory, produces a 128-dimensional vector representation (Fig. \ref{fig11}). Rao et al. \cite{rao2017learning} aggregated raw video frames directly by combining the idea of metric learning and adversarial learning. For dealing with bad frames, Rao et al. \cite{rao2017attention} discarded the bad frames by treating this operation as a Markov decision process and trained the attention model through a deep reinforcement learning framework. Ding et al. \cite{ding2017trunk} artificially blurred clear images for training to learn blur-robust face representations. Parchami et al. \cite{parchami2017using} used a CNN to reconstruct a lower-quality video into a high-quality face.

\subsection{Face Recognition in Industry}

\subsubsection{3D Face Recognition}\label{3D}

3D FR has inherent advantages over 2D methods, but 3D deep FR is not well developed due to the lack of large annotated 3D data. To enlarge 3D training datasets, most works use the methods of ``one-to-many augmentation'' to synthesize 3D faces. However, the effective methods for extracting deep features of 3D faces remain to be explored. Kim et al. \cite{kim2017deep} fine-tuned a 2D CNN with a small amount of 3D scans for 3D FR. Zulqarnain et al. \cite{zulqarnain2018learning} used a three-channel (corresponding to depth, azimuth and elevation angles of the normal vector) image as input and minimized the average prediction log-loss. Zhang et al. \cite{zhang2016research} first selected 30 feature points from the Candide-3 face model to characterize faces, then conducted the unsupervised pretraining of face depth data, and finally performed the supervised fine-tuning.

\subsubsection{Partial Face Recognition}\label{partial}

Partial FR, in which only arbitrary-size face patches are presented, has become an emerging problem with increasing requirements of identification from CCTV cameras and embedded vision systems in mobile devices, robots and smart home facilities. He et al. \cite{He2016Multiscale} divided the aligned face image into several multi-scale patches, and the dissimilarity between two partial face images is calculated as the weighted L2 distance between corresponding patches. Dynamic feature matching (DFM) \cite{He_2018_CVPR} utilized a sliding window of the same size as the probe feature maps to decompose the gallery feature maps into several gallery sub-feature maps, and the similarity-guided constraint imposed on sparse representation classification (SRC) provides an alignment-free
matching.

\subsubsection{Face Recognition for Mobile Devices}

With the emergence of mobile phones, tablets and augmented reality, FR has been applied in mobile devices. Due to computational limitations, the recognition tasks in these devices need to be carried out in a light but timely fashion.  MobiFace \cite{duong2018mobiface} required efficient memory and low cost operators by adopting fast downsampling and bottleneck residual block, and achieves
99.7\% on LFW database and 91.3\% on Megaface database. Tadmor et al. \cite{tadmor2016learning} proposed a multibatch method that first generates signatures for a minibatch of $k$ face images and then constructs an unbiased estimate of the full gradient by relying on all $k^2-k$ pairs from the minibatch. As mentioned in Section \ref{section}, light-weight deep networks \cite{iandola2016squeezenet,howard2017mobilenets,chollet2017xception,zhang2018shufflenet} perform excellently in the fundamental tasks of image classification and deserve further attention in FR tasks. Moreover, some well-known compressed networks such as Pruning \cite{han2015deep,han2015learning,liu2017learning}, BinaryNets \cite{courbariaux2016binarized,hubara2016binarized,rastegari2016xnor,courbariaux2015binaryconnect}, Mimic Networks \cite{li2017mimicking,wei2018quantization}, also have potential to be introduced into FR.

\subsubsection{Face Anti-attack} \label{anti-spoofing}

With the success of FR techniques, various types of attacks, such as face spoofing and adversarial perturbations, are becoming large threats. 
Face spoofing involves presenting a fake face to the biometric sensor using a printed photograph, worn mask, or even an image displayed on another electronic device. In order to defense this type of attack, several methods are proposed \cite{atoum2017face,yang2014learn,xu2015learning,li2016original,patel2016cross,Jourabloo_2018_ECCV,shao2019joint,shao2017deep}. Atoum et al. \cite{atoum2017face} proposed a novel two-stream CNN in which the local features discriminate the spoof patches that are independent of the spatial face areas, and holistic depth maps ensure that the input live sample has a face-like depth. Yang et al. \cite{yang2014learn} trained a CNN using both a single frame and multiple frames with five scales as input, and using the live/spoof label  as the output. Taken the sequence of video frames as input, Xu et al. \cite{xu2015learning} applied LSTM units on top of CNN to obtain end-to-end features to recognize spoofing faces which leveraged the local and dense property from convolution operation and learned the temporal structure using LSTM units. Li et al. \cite{li2016original} and Patel et al. \cite{patel2016cross} fine-tuned their networks from a pretrained model by training sets of real and fake images. Jourabloo et al. \cite{Jourabloo_2018_ECCV} proposed to inversely decompose a spoof face into the live face and the spoof noise pattern. 
Adversarial perturbation is the other type of attack which can be defined as the addition of a minimal vector $r$ such that with addition of this vector into the input image $x$, i.e. $(x+r)$, the deep learning models misclassifies the input while people will not. Recently, more and more work has begun to focus on solving this perturbation of FR. Goswami et al. \cite{goswami2018unravelling} proposed to detect adversarial samples by characterizing abnormal filter response behavior in the hidden layers and increase the network's robustness by removing the most problematic filters. Goel et al. \cite{goel2018smartbox} provided an open source implementation of adversarial detection and mitigation algorithms. 
Despite of progresses of anti-attack algorithms, attack methods are updated as well and remind us the need to further increase security and robustness in FR systems, for example, Mai et al. \cite{mai2018reconstruction} proposed a neighborly de-convolutional neural network (NbNet) to reconstruct a fake face using the stolen deep templates.

\subsubsection{Debiasing face recognition}

As described in Section \ref{dataset_ref}, existing datasets are highly biased in terms of the distribution of demographic cohorts, which may dramatically impact the fairness of deep models. To address this issue, there are some works that seek to introduce fairness into face recognition and mitigate demographic bias, e,g. unbalanced-training \cite{wang2020mitigating}, attribute removal \cite{alvi2018turning,mirjalili2018gender,othman2014privacy} and domain adaptation \cite{wang2019racial,guo2020learning,Kan2015Bi}. 1) Unbalanced-training methods mitigate the bias via model regularization, taking into consideration of the fairness goal in the overall model objective function. For example, RL-RBN \cite{wang2020mitigating} formulated the process of finding the optimal margins for non-Caucasians as a Markov decision process and employed deep Q-learning to learn policies based on large margin loss. 2) Attribute removal methods confound or remove demographic information of faces to learn attribute-invariant representations. For example, Alvi et al. \cite{alvi2018turning} applied a confusion loss to make a classifier fail to distinguish attributes of examples so that multiple spurious variations are removed from the feature representation. SensitiveNets \cite{morales2019sensitivenets} proposed to introduce sensitive information into triplet loss. They minimized the sensitive information, while maintaining distances between positive and negative embeddings. 3) Domain adaptation methods propose to investigate data bias problem from a domain adaptation point of view and attempt to design domain-invariant feature representations to mitigate bias across domains. IMAN \cite{wang2019racial}  
simultaneously aligned global distribution to decrease race gap at domain-level, and learned the discriminative target representations at cluster level. Kan \cite{Kan2015Bi} directly converted the Caucasian data to non-Caucasian domain in the image space with the help of sparse reconstruction coefficients learnt in the common subspace.

\section{Technical Challenges}

In this paper, we provide a comprehensive survey of deep FR from both data and algorithm aspects. For algorithms, mainstream and special network architectures are presented. Meanwhile, we categorize loss functions into Euclidean-distance-based loss, angular/cosine-margin-based loss and variable softmax loss. For data, we summarize some commonly used datasets. Moreover, the methods of face processing are introduced and categorized as ``one-to-many augmentation'' and ``many-to-one normalization''. Finally, the special scenes of deep FR, including video FR, 3D FR and cross-age FR, are briefly introduced.

Taking advantage of big annotated data and revolutionary deep learning techniques, deep FR has dramatically improved the SOTA performance and fostered successful real-world applications. With the practical and commercial use of this technology, many ideal assumptions of academic research were broken, and more and more real-world issues are emerging. To the best our knowledge, major technical challenges include the following aspects.

\begin{itemize}
\item \textbf{Security issues.} Presentation attack \cite{ramachandra2017presentation}, adversarial attack \cite{goswami2018unravelling,goel2018smartbox,zhong2020towards}, template attack \cite{mai2017reconstruction} and digital manipulation attack \cite{dang2020detection,Rossler_2019_ICCV} are developing to threaten the security of deep face recognition systems. 1) Presentation attack with 3D silicone mask, which exhibits skin-like appearance and facial motion, challenges current anti-sproofing methods \cite{manjani2017detecting}. 2) Although adversarial perturbation detection and mitigation methods are recently proposed \cite{goswami2018unravelling}\cite{goel2018smartbox}, the root cause of adversarial vulnerability is unclear and thus new types of adversarial attacks are still upgraded continuously \cite{sharif2017adversarial,sharif2016accessorize}. 3) The stolen deep feature template can be used to recover its facial appearance, and how to generate cancelable template without loss of accuracy is another important issue. 4) Digital manipulation attack, made feasible by GANs, can generate entirely or partially modified photorealistic faces by expression swap, identity swap, attribute manipulation and entire face synthesis, which remains a main challenge for the security of deep FR.
\end{itemize}

\begin{itemize}
\item \textbf{Privacy-preserving face recognition.} With the leakage of biological data, privacy concerns are raising nowadays. Facial images can predict not only demographic information such as gender, age, or race, but even the genetic information \cite{Gurovich2019identifying}. Recently, the pioneer works such as Semi-Adversarial Networks \cite{mirjalili2017soft,mirjalili2018semi,mirjalili2018gender} have explored to generate a recognizable biometric templates that can hidden some of the private information presented in the facial images. Further research on the principles of visual cryptography, signal mixing and image perturbation to protect users' privacy on stored face templates are essential for addressing public concern on privacy.
\end{itemize}

\begin{itemize}
\item \textbf{Understanding deep face recognition.} Deep face recognition systems are now believed to surpass human performance in most scenarios \cite{phillips2018face}. There are also some interesting attempts to apply deep models to assist human operators for face verification \cite{deng2017fine}\cite{phillips2018face}. Despite this progress, many fundamental questions are still open, such as what is the ``identity capacity'' of a deep representation \cite{gong2017capacity}? Why deep neural networks, rather than humans, are easily fooled by adversarial samples? While bigger and bigger training dataset by itself cannot solve this problem, deeper understanding on these questions may help us to build robust applications in real world. Recently, a new benchmark called TALFW has been proposed to explore this issue \cite{zhong2017toward}.
\end{itemize}

\begin{itemize}
\item \textbf{Remaining challenges defined by non-saturated benchmark datasets.} Three current major datasets, namely, MegaFace \cite{kemelmacher2016megaface,nech2017level} , MS-Celeb-1M \cite{guo2016ms} and IJB-A/B/C \cite{klare2015pushing,Whitelam2017IARPA,maze2018iarpa}, are corresponding to large-scale FR with a very large number of candidates, low/one-shot FR and large pose-variance FR which will be the focus of research in the future. Although the SOTA algorithms can be over 99.9 percent accurate on LFW \cite{huang2007labeled} and Megaface \cite{kemelmacher2016megaface,nech2017level} databases, fundamental challenges such as matching faces cross ages \cite{CPLFW}, poses \cite{zheng2017cross}, sensors, or styles still remain. For both datasets and algorithms, it is necessary to measure and address the racial/gender/age biases of deep FR in future research.
\end{itemize}

\begin{itemize}
\item \textbf{Ubiquitous face recognition across applications and scenes.} Deep face recognition has been successfully applied on many user-cooperated applications, but the ubiquitous recognition applications in everywhere are still an ambitious goal. In practice, it is difficult to collect and label sufficient samples for innumerable scenes in real world. One promising solution is to first learn a general model and then transfer it to an application-specific scene. While deep domain adaptation \cite{wang2018deep} has recently been applied to reduce the algorithm bias on different scenes \cite{Luo2018Adaptation}, different races \cite{wang2019racial}, general solution to transfer face recognition is largely open. 
\end{itemize}

\begin{itemize}
\item \textbf{Pursuit of extreme accuracy and efficiency.} Many killer-applications, such as watch-list surveillance or financial identity verification, require high matching accuracy at very low alarm rate, e.g. $10^{-9}$. It is still a big challenge even with deep learning on massive training data. Meanwhile, deploying deep face recognition on mobile devices pursues the minimum size of feature representation and compressed deep network. It is of great significance for both industry and academic to explore this extreme face-recognition performance beyond human imagination. It is also exciting to constantly push the performance limits of the algorithm after it has already surpassed human.
\end{itemize}

\begin{itemize}
\item \textbf{Fusion issues.} Face recognition by itself is far from sufficient to solve all biometric and forensic tasks, such as distinguishing identical twins and matching faces before and after surgery \cite{singh2010plastic}. A reliable solution is to consolidate multiple sources of biometric evidence \cite{ross2004multimodal}. These sources of information may correspond to different biometric traits (e.g., face + hand \cite{ross2005feature}), sensors (e.g., 2D + 3D face cameras), feature extraction and matching techniques, or instances (e.g., a face sequence of various poses). It is beneficial for face biometric and forensic applications to perform information fusion at the data level, feature level, score level, rank level, and decision level \cite{ross2003information}.
\end{itemize}


\section{Acknowledgments}

This work was partially supported by National Key R\&D Program of China (2019YFB1406504) and BUPT Excellent Ph.D. Students Foundation CX2020207.

{\footnotesize
\bibliographystyle{IEEEtran}
\bibliography{egbib}
}

\end{document}